\documentclass{article}

    \PassOptionsToPackage{numbers, compress}{natbib}

\usepackage[final]{neurips_2021}

\usepackage[utf8]{inputenc} 
\usepackage[T1]{fontenc}    
\usepackage{hyperref}       
\usepackage{url}            
\usepackage{booktabs}       
\usepackage{amsfonts}       
\usepackage{nicefrac}       
\usepackage{microtype}      
\usepackage{xcolor}         

\usepackage{amsmath}
\usepackage{dsfont}

\usepackage{placeins}

\usepackage{graphicx}
\usepackage{wrapfig}
\usepackage[ruled]{algorithm2e}
\usepackage{multirow}
\usepackage{mathtools}

\SetArgSty{textnormal}

\DeclareMathOperator*{\argmax}{arg\,max}
\DeclareMathOperator*{\argmin}{arg\,min}
\newcommand{\jac}{\mathbf J}
\newcommand{\R}{\mathbb R}

\usepackage{subcaption}

\title{Rectangular Flows for Manifold Learning}

\author{%
  Anthony L. Caterini\thanks{Authors contributed equally.} \\
  University of Oxford \& Layer 6 AI\\
  \texttt{anthony@layer6.ai}
   \And
   Gabriel Loaiza-Ganem$^*$ \\
   Layer 6 AI \\
   \texttt{gabriel@layer6.ai}
   \AND
   Geoff Pleiss \\
   Columbia University \\
   \texttt{gmp2162@columbia.edu}
   \And
   John P. Cunningham \\
   Columbia University \\
   \texttt{jpc2181@columbia.edu}
}

\begin{document}

\maketitle

\begin{abstract}
  Normalizing flows are invertible neural networks with tractable change-of-volume terms, which allow optimization of their parameters to be efficiently performed via maximum likelihood.
  However, data of interest are typically assumed to live in some (often unknown) low-dimensional manifold embedded in a high-dimensional ambient space.
  The result is a modelling mismatch since -- by construction -- the invertibility requirement implies high-dimensional support of the learned distribution. Injective flows, mappings from low- to high-dimensional spaces, aim to fix this discrepancy by learning distributions on manifolds, but the resulting volume-change term becomes more challenging to evaluate.
  Current approaches either avoid computing this term entirely using various heuristics, or assume the manifold is known beforehand and therefore are not widely applicable.
  Instead, we propose two methods to tractably calculate the gradient of this term with respect to the parameters of the model, relying on careful use of automatic differentiation and techniques from numerical linear algebra.
  Both approaches perform end-to-end nonlinear manifold learning and density estimation for data projected onto this manifold.
We study the trade-offs between our proposed methods, empirically verify that we outperform approaches ignoring the volume-change term by more accurately learning manifolds and the corresponding distributions on them, and show promising results on out-of-distribution detection. Our code is available at \url{https://github.com/layer6ai-labs/rectangular-flows}.
\end{abstract}

\section{Introduction}

In recent years, Normalizing Flows (NFs) have become a staple of generative modelling, being widely used for density estimation \citep{dinh2014nice, dinh2016density, papamakarios2017masked, kingma2018glow, durkan2019neural}, variational inference \citep{rezende2015variational, kingma2016improved}, maximum entropy modelling \citep{loaiza2017maximum}, and more \citep{papamakarios2019normalizing, kobyzev2020normalizing}.
In density estimation, we typically have access to a set of points living in some high-dimensional space $\R^D$.
NFs model the corresponding data-generating distribution as the pushforward of a simple distribution on $\R^D$ -- often a Gaussian -- through a smooth bijective mapping.
Clever construction of these bijections allows for tractable density evaluation and thus maximum likelihood estimation of the parameters.
However, as an immediate consequence of this choice, the learned distribution has support homeomorphic to $\R^D$;
in particular, the resulting distribution is supported on a set of dimension $D$.
This is not a realistic assumption in practice -- especially for density estimation -- as it directly contradicts the manifold hypothesis \citep{bengio2013representation} which states that high-dimensional data lives on a lower-dimensional manifold embedded in ambient space.

A natural idea to circumvent this misspecification is to consider \emph{injective} instead of bijective flows, which now push forward a random variable on $\R^d$ with $d < D$ to obtain a distribution on some $d$-dimensional manifold embedded in $\R^D$.
These mappings admit a change-of-variable formula bearing resemblance to that of bijective flows, but unfortunately the volume-change term becomes computationally prohibitive, which then impacts the tractability of maximum likelihood.
While there have been recent efforts towards training flows where the resulting distribution is supported on a low-dimensional manifold \citep{gemici2016normalizing, rezende2020normalizing, brehmer2020flows, kumar2020regularized, mathieu2020riemannian, cunningham2020normalizing}, these approaches either assume that the manifold is known beforehand or propose various heuristics to avoid the change-of-variable computation.
Both of these are undesirable, because, while we should expect most high-dimensional data of interest to exhibit low-dimensional structure, this structure is almost always unknown.
On the other hand, we argue that avoiding the volume-change term may result in learning a manifold to which it is difficult to properly assign density, and this approach further results in methods which do not take advantage of density evaluation, undermining the main motivation for using NFs in the first place.

We show that density estimation for injective flows based on maximum likelihood can be made tractable.
By carefully leveraging forward- and backward-mode automatic differentiation \citep{baydin2018automatic}, we propose two methods that allow backpropagating through the volume term arising from the injective change-of-variable formula.
The first method involves exact evaluation of this term and its gradient which incurs a higher memory cost;
the second uses conjugate gradients \citep{nocedal2006numerical} and Hutchinson's trace estimator \citep{hutchinson1989stochastic} to obtain unbiased stochastic gradient estimates.
Unlike previous work, our methods do not need the data manifold to be specified beforehand, but instead simultaneously estimate this manifold along with the distribution on it end-to-end, thus enabling maximum likelihood training to occur.
To the best of our knowledge, ours are the first methods to scale backpropagation through the injective volume-change term to ambient dimensions $D$ close to $3{,}000$. 
We study the trade-off between memory and variance introduced by our methods and show empirical improvements over injective flow baselines for density estimation. We also show that injective flows obtain state-of-the-art performance for likelihood-based Out-of-Distribution (OoD) detection, assigning higher likelihoods to Fashion-MNIST (FMNIST) \citep{xiao2017fashion} than to MNIST \citep{lechun1998mnist} with a model trained on the former.

\section{Background}\label{sec:background}

\subsection{Square Normalizing Flows}

A normalizing flow \citep{rezende2015variational, dinh2016density} is a diffeomorphism $\tilde{f}_\theta:\R^D \rightarrow \R^D$ parametrized by $\theta$, that is, a differentiable bijection with differentiable inverse.
Starting with a random variable $Z \sim p_Z$ for a simple density $p_Z$ supported on $\R^D$, e.g.\ a standard Gaussian, the change-of-variable formula states that the random variable $X := \tilde{f}_\theta(Z)$ has density $p_X$ on $\R^D$ given by:
\begin{equation}\label{eq:square_density}
    p_X(x) = p_Z\left(\tilde{f}^{-1}_\theta(x)\right) \left| \det \jac \left[\tilde{f}_\theta\right] \left(\tilde{f}_\theta^{-1}(x)\right)\right|^{-1},
\end{equation}
where $\jac [\cdot]$ is the differentiation operator, so that $\jac[\tilde{f}_\theta](\tilde{f}_\theta^{-1}(x)) \in \mathbb{R}^{D \times D}$ is the Jacobian of $\tilde{f}_\theta$ (with respect to the inputs and not $\theta$) evaluated at $\tilde{f}_\theta^{-1}(x)$.
We refer to this now standard setup as \textit{square flows} since the Jacobian is a square matrix.
The change-of-variable formula is often written in terms of the Jacobian of $\tilde{f}^{-1}_\theta$, but we use the form of \eqref{eq:square_density} as it is more applicable for the next section.
NFs are typically constructed in such a way that not only ensures bijectivity, but also so that the Jacobian determinant in \eqref{eq:square_density} can be efficiently evaluated.
When provided with a dataset $\{x_i\}_{i=1}^n \subset \R^D$, an NF models its generating distribution as the pushforward of $p_Z$ through $\tilde{f}_\theta$, and thus the parameters can be estimated via maximum likelhood as $\theta^* \coloneqq \argmax_\theta \sum_{i=1}^n \log p_X(x_i)$. 

\subsection{Rectangular Normalizing Flows}\label{subsec:rectangular}

As previously mentioned, square NFs unrealistically result in the learned density $p_X$ having $D$-dimensional support. We follow the injective flow construction of \citet{brehmer2020flows}, where a smooth and injective mapping $g_\phi: \R^d \rightarrow \R^D$ with $d < D$ is constructed.
In this setting, $Z \in \R^d$ is the low-dimensional variable used to model the data as $X:=g_\phi(Z)$.
A well-known result from differential geometry \citep{krantz2008geometric} provides an applicable change-of-variable formula:
\begin{equation}\label{eq:rectangular_density}
    p_X(x) = p_Z\left(g_\phi^{-1}(x)\right)\left| \det \jac[g_\phi]^\top\left(g_\phi^{-1}(x)\right) \jac [g_\phi]\left(g_\phi^{-1}(x)\right)\right|^{-1/2} \mathds{1}(x \in \mathcal{M}_\phi),
\end{equation}
where $\mathcal{M}_\phi \coloneqq \{g_\phi(z): z\in \mathbb{R}^d\}$.
The Jacobian-transpose-Jacobian determinant now characterizes the change in volume from $Z$ to $X$.
We make several relevant observations: $(i)$ The Jacobian matrix $\jac[g_\phi](g_\phi^{-1}(x)) \in \mathbb{R}^{D \times d}$ is no longer a square matrix, and we thus refer to these flows as \textit{rectangular}. $(ii)$ Note that $g_\phi^{-1}:\mathcal{M}_\phi \rightarrow \R^d$ is only properly defined on $\mathcal{M}_\phi$ and not $\R^D$, and $p_X$ is now supported on the $d$-dimensional manifold $\mathcal{M}_\phi$.
$(iii)$ We write the indicator $\mathds{1}(x \in \mathcal{M}_\phi)$ explicitly to highlight the fact that this density is \textit{not} a density with respect to the Lebesgue measure;
rather, the dominating measure is a Riemannian measure on the manifold $\mathcal{M}_\phi$ \citep{pennec2006intrinsic}.
$(iv)$ One can clearly verify as a sanity check that when $d=D$, equation \eqref{eq:rectangular_density} reduces to \eqref{eq:square_density}.

Since data points $x$ will almost surely not lie exactly on $\mathcal{M}_\phi$, we use a left inverse $g_\phi^\dagger:\R^D \rightarrow \R^d$ in place of $g_\phi^{-1}$ such that $g_\phi^\dagger(g_\phi(z))=z$ for all $z \in \R^d$, which exists because $g_\phi$ is injective.
This is properly defined on $\R^D$, unlike $g_\phi^{-1}$ which only exists over $\mathcal{M}_\phi$.
Equation \eqref{eq:rectangular_density} then becomes:
\begin{equation}\label{eq:rectangular_density_proj}
    p_X(x) = p_Z\left(g_\phi^\dagger(x)\right)\left| \det \jac [g_\phi]^\top\left(g_\phi^\dagger(x)\right) \jac [g_\phi]\left(g_\phi^\dagger(x)\right)\right|^{-1/2}.
\end{equation}
Note that \eqref{eq:rectangular_density_proj} is equivalent to projecting $x$ onto $\mathcal{M}_\phi$ as $x \leftarrow g_\phi(g_\phi^\dagger(x))$, and then evaluating the density from \eqref{eq:rectangular_density} at the projected point.

Now, $g_\phi$ is injectively constructed as follows:
\begin{equation}\label{eq:injective_flow}
    g_\phi = \tilde{f}_\theta \circ \texttt{pad} \circ h_\eta \hspace{20pt}\text{and}\hspace{20pt} g_\phi^\dagger = h_\eta^{-1} \circ \texttt{pad}^\dagger \circ \tilde{f}_\theta^{-1},
\end{equation}
where $\tilde{f}_\theta : \R^D \rightarrow \R^D$ and $h_\eta : \R^d \rightarrow \R^d$ are both square flows, $\phi
\coloneqq (\theta, \eta)$, and $\texttt{pad}:\mathbb{R}^d \rightarrow \mathbb{R}^D$ and $\texttt{pad}^\dagger:\mathbb{R}^D \rightarrow \mathbb{R}^d$ are defined as $\texttt{pad}(z) = (z, \mathbf{0})$  and $\texttt{pad}^\dagger(z, z') = z$, where $\mathbf{0}, z' \in \mathbb{R}^{D-d}$.
Now, $\mathcal{M}_\phi$ depends only on $\theta$ and not $\eta$, so we write it as $\mathcal{M}_\theta$ from now on.
Applying \eqref{eq:rectangular_density_proj} yields:
\begin{equation}\label{eq:rectangular_density_2}
    p_X(x) = p_Z\left(g_\phi^\dagger(x)\right)\left|\det \jac [h_\eta]\left(g_\phi^\dagger(x)\right)\right|^{-1}\left| \det \jac [f_\theta]^\top\left(f_\theta^\dagger(x)\right) \jac [f_\theta]\left(f_\theta^\dagger(x)\right)\right|^{-1/2},
\end{equation}
where $f_\theta = \tilde{f}_\theta \circ \texttt{pad}$ and $f_\theta^\dagger = \texttt{pad}^\dagger \circ \tilde{f}_\theta^{-1}$.
We include a derivation of \eqref{eq:rectangular_density_2} in \autoref{app:change}, along with a note on why injective transformations cannot be stacked as naturally as bijective ones.

Evaluating likelihoods is seemingly intractable since constructing flows with a closed-form volume-change term is significantly more challenging than in the square case, even if the relevant matrix is now $d \times d$ instead of $D \times D$.
\citet{brehmer2020flows} thus propose a two-step training procedure to promote tractability wherein $f_\theta$ and $h_\eta$ are trained separately.
After observing that there is no term encouraging $x \in \mathcal{M}_\theta$, and that $x \in \mathcal{M}_\theta \iff x=g_\phi(g_\phi^\dagger(x)) \iff x = f_\theta(f_\theta^\dagger (x))$, they decide to simply train $f_\theta$ by minimizing the reconstruction error to encourage the observed data to lie on $\mathcal{M}_\theta$:
\begin{equation}
    \theta^* = \argmin_\theta \displaystyle \sum_{i=1}^n \left|\left|x_i - f_\theta\left(f_\theta^\dagger(x_i)\right) \right|\right|_2^2.
\end{equation}
Note that the above requires computing both $f_\theta$ and $f_\theta^\dagger$, so that $\tilde{f}_\theta$ should be chosen as a flow allowing fast evaluation of both $\tilde{f}_\theta$ and $\tilde{f}_\theta^{-1}$.
Architectures such as the Real NVP \citep{dinh2016density} or follow-up work \citep{kingma2018glow,durkan2019neural} are thus natural choices for $\tilde{f}_\theta$, while architectures with an autoregressive component \citep{papamakarios2017masked, kingma2016improved} should be avoided.
Then, since $h_\eta$ does not appear in the challenging determinant term in \eqref{eq:rectangular_density_2}, $h_\eta$ can be chosen as any normalizing flow, and optimization -- for a fixed $\theta$ -- can be tractably achieved by maximum likelihood over the lower-dimensional space:
\begin{equation}
    \eta^* = \argmax_\eta \sum_{i=1}^n \left\{ \log p_Z\left(g_\phi^\dagger(x_i)\right) - \log \left|\det \jac [h_\eta]\left(g_\phi^\dagger(x_i)\right)\right|\right\}.
\end{equation}
In practice, gradient steps in $\theta$ and $\eta$ are alternated.
This entire procedure circumvents evaluation of the Jacobian-transpose-Jacobian determinant term in \eqref{eq:rectangular_density_2}, but as we show in \autoref{sec:related}, avoiding this term by separately learning the manifold and the density on it comes with its downsides.
We then show how to tractably estimate this term in \autoref{sec:method}.
\section{Related Work and Motivation}\label{sec:related}

\paragraph{Low-dimensional and topological pathologies} The mismatch between the dimension of the modelled support and that of the data-generating distribution has been observed throughout the literature in various ways.
\citet{dai2019diagnosing} show, in the context of variational autoencoders \citep{kingma2013auto}, that using flexible distributional approximators supported on $\R^D$ to model data living in a low-dimensional manifold results in pathological behavior where the manifold itself is learned, but not the distribution on it.
\citet{cornish2020relaxing} demonstrate the drawbacks of normalizing flows for estimating the density of topologically-complex data, and provide a new numerically stable method for learning NFs when the support is not homeomorphic to $\R^D$.
However, this approach still models the support as being $D$-dimensional.
\citet{behrmann2021understanding} show instabilities associated with NFs -- particularly a lack of numerical invertibility, as also explained theoretically by \citet{cornish2020relaxing}.
This is not too surprising, as attempting to learn a smooth invertible function mapping $\R^D$ to some low-dimensional manifold is an intrinsically ill-posed problem.
This body of work strongly motivates the development of models whose support has matching topology -- including dimension -- to that of the true data distribution.

\paragraph{Manifold flows} A challenge to overcome for obtaining NFs on manifolds is the Jacobian-transpose-Jacobian determinant computation.
Current approaches for NFs on manifolds approach this challenge in one of two ways.
The first assumes the manifold is known beforehand \citep{gemici2016normalizing, rezende2020normalizing, mathieu2020riemannian}, severely limiting applicability to low-dimensional data where the true manifold can realistically be known.
The second group circumvents the computation of the Jacobian-transpose-Jacobian entirely through various heuristics.
\citet{kumar2020regularized} use a potentially loose lower bound of the log-likelihood, and do not explicitly enforce injectivity, resulting in a method for which the change-of-variables almost surely does not hold. 
\citet{cunningham2020normalizing} propose to convolve the manifold distribution with Gaussian noise, which results in the model having high-dimensional support.
Finally, \citet{brehmer2020flows} propose the method we described in \autoref{subsec:rectangular}, where manifold learning and density estimation are done separately in order to avoid the log determinant computation.
Concurrently to our work, \citet{ross2021tractable} proposed a rectangular flow construction which sacrifices some expressiveness but allows for exact likelihood evaluation.

\paragraph{Why optimize the volume-change term?} Learning $f_\theta$ and $h_\eta$ separately without the Jacobian of $f_\theta$ is concerning: 
even if $f_\theta$ maps to the correct manifold, it might unnecessarily expand and contract volume in such a way that makes correctly learning $h_\eta$ much more difficult than it needs to be.
Looking ahead to our experiments, \autoref{fig:circle} exemplifies this issue: the \textbf{top-middle} panel shows the ground truth density on a 1-dimensional circle in $\mathbb{R}^2$, and the \textbf{top-right} panel shows the distribution recovered by the two-step method of \citet{brehmer2020flows}.
We can see that, while the manifold is correctly recovered, the distribution on it is not.
The \textbf{bottom-right} panel shows the speed at which $f_{\theta^*}$ maps $\R$ to $\mathcal{M}_{\theta^*}$: the top of the circle, which should have large densities, also has high speeds.
Indeed, there is nothing in the objective discouraging $f_\theta$ to learn this behaviour, which implies that the corresponding low-dimensional distribution must be concentrated in a small region and thus making it harder to learn.
The \textbf{bottom-middle} panel confirms this explanation: the learned low-dimensional distribution (dark red) does not match what it should (i.e.\ the distribution of $\{f^{\dagger}_{\theta^*}(x_i)\}_{i=1}^n$, in light red).
This failure could have been avoided by learning the manifold in a density-aware fashion by including the Jacobian-transpose-Jacobian determinant in the objective.

\section{Maximum Likelihood for Rectangular Flows: Taming the Gradient}\label{sec:method}

\subsection{Our Optimization Objective}

We have argued that including the Jacobian-transpose-Jacobian in the optimization objective is sensible.
However, as we previously mentioned, \eqref{eq:rectangular_density_2} corresponds to the density of the projection of $x$ onto $\mathcal{M}_\theta$.
Thus, simply optimizing the likelihood would not result in learning $\mathcal{M}_\theta$ in such a way that observed data lies on it, only encouraging \emph{projected} data points to have high likelihood.
We thus maximize the log-likelihood subject to the constraint that the reconstruction error should be smaller than some threshold, i.e.\ $\phi^* = \argmax_\phi \sum_{i=1}^n \log p_X(x_i)$ subject to $\sum_{i=1}^n ||x_i - f_\theta(f_\theta^\dagger (x_i))||_2^2 \leq \kappa$.
In practice, we use the KKT conditions \citep{karush1939minima, kuhn1951w} and maximize the Lagrangian \citep{bertsekas2014constrained} instead:
\begin{align}
    \phi^*\!  = \argmax_\phi \sum_{i=1}^n & \left\{ \log p_Z\!\left(g_\phi^\dagger(x_i)\right) - \log \left|\det \jac [h_\eta]\!\left(g_\phi^\dagger(x_i)\right)\right| - \dfrac{1}{2}\log  \det J_{\theta}^\top(x_i) J_{\theta}(x_i)\right. \label{eq:reg_obj} \\
    & \left. \quad - \beta \left|\left|x_i - f_\theta\left(f_\theta^\dagger(x_i)\right) \right|\right|_2^2\right\} \nonumber,
\end{align}
where we treat $\beta > 0$ as a hyperparameter rather than $\kappa$, and denote $\jac [f_\theta](f_\theta^\dagger(x_i))$ as $J_\theta(x_i)$ for simplicity.
We have dropped the absolute value since $J_\theta^\top(x_i)J_\theta(x_i)$ is always symmetric positive definite, since $J_\theta(x_i)$ has full rank by injectivity of $f_\theta$.
We now make a technical but relevant observation about our objective: since our likelihoods are Radon-Nikodym derivatives with respect to a Riemannian measure on $\mathcal{M}_\theta$, different values of $\theta$ will result in different supports and dominating measures.
One should thus be careful to compare likelihoods for models with different values of $\theta$.
However, thanks to the smoothness of the objective over $\theta$, we should expect likelihoods for values of $\theta$ which are ``close enough'' to be comparable for practical purposes.
In other words, comparisons remain reasonable locally, and the gradient of the volume-change term should contain relevant information to learn $\mathcal M_\theta$ in such a way that also facilitates learning $h_\eta$ on the pulled-back dataset $\{f_\theta^\dagger(x_i)\}_{i=1}^n$.

\subsection{Optimizing our Objective: Stochastic Gradients}

Note that all the terms in \eqref{eq:reg_obj} are straightforward to evaluate and backpropagate through except for the third one; in this section we show how to obtain unbiased stochastic estimates of its gradient.
In what follows we drop the dependence of the Jacobian on $x_i$ from our notation and write $J_\theta$, with the understanding that the end computation will be parallelized over a batch of $x_i$s.
We assume access to an efficient matrix-vector product routine, i.e.\ computing $J_\theta^\top J_\theta \epsilon$ can be quickly achieved for any $\epsilon \in \mathbb{R}^d$.
We elaborate on how we obtain these matrix-vector products in the next section.
It is a well known fact from matrix calculus \citep{Petersen2008} that:
\begin{equation}
    \dfrac{\partial}{\partial \theta_j} \log \det J_\theta^\top J_\theta = \mathrm{tr}\left((J_\theta^\top J_\theta)^{-1}\dfrac{\partial}{\partial \theta_j} J_\theta^\top J_\theta\right),
\end{equation}
where $\mathrm{tr}$ denotes the trace operator and $\theta_j$ is the $j$-th element of $\theta$.
Next, we can use Hutchinson's trace estimator \citep{hutchinson1989stochastic}, which states that for any matrix $M \in \mathbb{R}^{d \times d}$, $\mathrm{tr}(M) = \mathbb{E}_\epsilon[\epsilon^\top M \epsilon]$ for any $\mathbb{R}^d$-valued random variable $\epsilon$ with zero mean and identity covariance matrix.
We can thus obtain an unbiased stochastic estimate of our gradient as:
\begin{equation}\label{eq:estimate_naive}
    \dfrac{\partial}{\partial \theta_j} \log \det J_\theta^\top J_\theta \approx \dfrac{1}{K} \displaystyle \sum_{k=1}^K \epsilon_k^\top (J_\theta^\top J_\theta)^{-1}\dfrac{\partial}{\partial \theta_j} J_\theta^\top J_\theta \epsilon_k,
\end{equation}
where $\epsilon_1,\dots,\epsilon_K$ are typically sampled either from standard Gaussian or Rademacher distributions.
Na\"ive computation of the above estimate remains intractable without explicitly constructing $J_\theta^\top J_\theta$.
Fortunately, the $J_\theta^\top J_\theta \epsilon$ terms can be trivially obtained using the given matrix-vector product routine, avoiding the construction of $J_\theta^\top J_\theta$, and then $\partial / \partial \theta_j J_\theta^\top J_\theta \epsilon$ follows by taking the gradient w.r.t.\ $\theta$.

There is however still the issue of computing $\epsilon^\top(J_\theta^\top J_\theta)^{-1} = [(J_\theta^\top J_\theta)^{-1}\epsilon]^\top$.
We use conjugate gradients (CG) \citep{nocedal2006numerical} in order to achieve this.
CG is an iterative method to solve problems of the form $Au=\epsilon$ for given $A \in \mathbb{R}^{d \times d}$ (in our case $A = J_\theta^\top J_\theta$) and $\epsilon \in \mathbb{R}^d$; we include the CG algorithm in Appendix \ref{app:cg} for completeness.
CG has several important properties.
First, it is known to recover the solution (assuming exact arithmetic) after at most $d$ steps, which means we can evaluate $A^{-1}\epsilon$.
The solution converges exponentially (in the number of iterations $\tau$) to the true value \citep{shewchuk1994introduction},
so often $\tau \ll d$ iterations are sufficient for accuracy to many decimal places.
In practice, if we can tolerate a certain amount of bias, we can further increase computational speed by stopping iterations early.
Second, CG only requires a method to compute matrix-vector products against $A$, and does not require access to $A$ itself.
One such product is performed at each iteration, and CG thus requires at most $d$ matrix-vector products,
though again in practice $\tau \ll d$ products usually suffice.
This results in $\mathcal O(\tau d^2 )$ solve complexity---less than the $\mathcal O(d^3)$ required by direct inversion methods.
We denote $A^{-1}\epsilon$ computed with conjugate gradients as $\texttt{CG}(A; \epsilon)$.
We can then compute the estimator from \eqref{eq:estimate_naive} as:
\begin{equation}
    \dfrac{\partial}{\partial \theta_j} \log \det J_\theta^\top J_\theta \approx \dfrac{1}{K} \displaystyle \sum_{k=1}^K \texttt{CG}\left(J_\theta^\top J_\theta; \epsilon_k\right)^\top \dfrac{\partial}{\partial \theta_j} J_\theta^\top J_\theta \epsilon_k.
\end{equation}
In practice, we implement this term  by noting that $\texttt{CG}(J_\theta^\top J_\theta; \epsilon)^\top \partial / \partial \theta_j J_\theta^\top J_\theta \epsilon = \partial / \partial \theta_j \texttt{stop\_gradient}(\texttt{CG}(J_\theta^\top J_\theta; \epsilon)^\top) J_\theta^\top J_\theta \epsilon$, thereby taking advantage of the $\texttt{stop\_gradient}$ operation from Automatic Differentiation (AD) libraries and allowing us to avoid implementing a custom backward pass. We thus compute the contribution of a point $x$ to the training objective as: 
\begin{align}\label{eq:final_obj}
    \log p_Z\left(g_\phi^\dagger(x)\right) & - \log \left|\det \jac [h_\eta]\left(g_\phi^\dagger(x)\right)\right| - \beta \left|\left|x - f_\theta\left(f^\dagger_\theta(x)\right)\right|\right|_2^2 \\
    & -\dfrac{1}{2K} \displaystyle \sum_{k=1}^K \texttt{stop\_gradient}\left(\texttt{CG}\left(J_\theta^\top J_\theta; \epsilon_k\right)^\top\right) J_\theta^\top J_\theta \epsilon_k \nonumber
\end{align}
which gives the correct gradient estimate when taking the derivative with respect to $\phi$.

\paragraph{Linear solvers for Jacobian terms} We note that linear solvers like CG have been used before to backpropagate through log determinant computations in the context of Gaussian processes \citep{gardner2018gpytorch}, and more recently for square NFs with flexible architectures which do not allow for straightforward Jacobian determinant computations \citep{huang2020convex, lu2021implicit}.
However, none of these methods require the Jacobian-transpose-Jacobian-vector product routine that we derive in the next section, and to the best of our knowledge, these techniques have not been previously applied for training rectangular NFs.
We also point out that recently \citet{oktay2020randomized} proposed a method to efficiently obtain stochastic estimates of $J_\theta \epsilon$.
While their method cannot be used as a drop-in replacement within our framework as it would result in a biased CG output, we believe this could be an interesting direction for future work.
Finally, we note that CG has recently been combined with the Russian roulette estimator \citep{kahn1955use} to avoid having to always iterate $d$ times while maintaining unbiasedness, again in the context of Gaussian processes \citep{potapczynski2021bias}.
We also leave the exploration of this estimator within our method for future work. 

\subsection{AD Considerations: The Exact Method and the Forward-Backward AD Trick}

In this section we derive the aforementioned routine for vector products against $J_\theta^\top J_\theta$, as well as an exact method that avoids the need for stochastic gradients (for a given $x$) at the price of increased memory requirements.
But first, let us ask: why are these methods needed in the first place?
There is work using power series to obtain stochastic estimates of log determinants \citep{han2015large, NEURIPS2019_5d0d5594}, and one might consider using them in our setting.
However, these series require knowledge of the singular values of $J_\theta^\top J_\theta$, to which we do not have access (constructing $J_\theta^\top J_\theta$ to obtain its singular values would defeat the purpose of using the power series in the first place), and we would thus not have a guarantee that the series are valid.
Additionally, they have to be truncated and thus result in biased estimators, and using Russian roulette estimators to avoid bias \citep{NEURIPS2019_5d0d5594} can result in infinite variance \citep{cornish2020relaxing}.
Finally, these series compute and backpropagate (w.r.t.\ $\theta$) through products of the form $\epsilon^\top (J_\theta^\top J_\theta)^m \epsilon$ for different values of $m$, which can easily require more matrix-vector products than our methods.
\citet{behrmann2019invertible} address some of the issues with power series approximations as the result of controlling Lipschitz constants, although their estimates remain biased and potentially expensive.

Having motivated our approach, we now use commonly-known properties of AD to derive it;  we briefly review these properties in \autoref{app:ad}, referring the reader to \citet{baydin2018automatic} for more detail.
First, we consider the problem of explicitly constructing $J_\theta$.
This construction can then be used to evaluate $J_\theta^\top J_\theta$ and exactly compute its log determinant either for log density evaluation of a trained model, or to backpropagate (with respect to $\theta$) through both the log determinant computation and the matrix construction, thus avoiding having to use stochastic gradients as in the previous section.
We refer to this procedure as the \emph{exact} method.
Na\"ively, one might try to explicitly construct $J_\theta$ using only backward-mode AD, which would require $D$ vector-Jacobian products (\texttt{vjp}s) of the form $v^\top J_\theta$ -- one per basis vector $v \in \R^D$ (and then stacking the resulting row vectors vertically).
A better way to explicitly construct $J_\theta$ is with forward-mode AD, which only requires $d$ Jacobian-vector products (\texttt{jvp}s) $J_\theta \epsilon$, again one per basis vector $\epsilon \in \mathbb{R}^d$ (and then stacking the resulting column vectors horizontally).
We use a custom implementation of forward-mode AD in the popular PyTorch \citep{NEURIPS2019_9015} library\footnote{PyTorch has a forward-mode AD implementation which relies on the ``double backward'' trick, which is known to be memory-inefficient.
See \url{https://j-towns.github.io/2017/06/12/A-new-trick.html} for a description.} for the exact method, as well as for the forward-backward AD trick described below.

We now explain how to combine forward- and backward-mode AD to obtain efficient matrix-vector products against $J_\theta^\top J_\theta$ in order to obtain the tractable gradient estimates from the previous section.
Note that $v:= J_\theta \epsilon$ can be computed with a single $\texttt{jvp}$ call, and then $J_\theta^\top J_\theta \epsilon = [v^\top J_\theta]^\top$ can be efficiently computed using only a $\texttt{vjp}$ call.
We refer to this way of computing matrix-vector products against $J_\theta^\top J_\theta$ as the \emph{forward-backward AD trick}.
We summarize both of our gradient estimators in Appendix \ref{app:summary}.
Note that \eqref{eq:final_obj} requires $K(\tau+1)$ such matrix-vector products, which is seemingly less efficient as it is potentially greater than the $d$ \texttt{jvp}s required by the exact method.
However, the stochastic method is much more memory-efficient than its exact counterpart when optimizing over $\theta$: of the $K(\tau+1)$ matrix-vector products needed to evaluate \eqref{eq:final_obj}, only $K$ require gradients with respect to $\theta$.
Thus only $K$ $\texttt{jvp}$s and $K$ $\texttt{vjp}$s, along with their intermediate steps, must be stored in memory over a training step. 
In contrast, the exact method requires gradients (w.r.t.\ $\theta$) for every one of its $d$ $\texttt{jvp}$ computations, which requires storing these computations along with their intermediate steps in memory.

Our proposed methods thus offer a memory vs.\ variance trade-off.
Increasing $K$ in the stochastic method results in larger memory requirements which imply longer training times, as the batch size must be set to a smaller value.
On the other hand, the larger the memory cost, the smaller the variance of the gradient.
This still holds true for the exact method, which results in exact gradients, at the cost of increased memory requirements (as long as $K \ll d$; if $K$ is large enough the stochastic method should never be used over the exact one).
Table \ref{table:ad_costs} summarizes this trade-off.

\begin{table}
\small
  \caption{Number of \texttt{jvp}s and \texttt{vjp}s (with respect to inputs) needed for forward and backward passes (with respect to $\theta$), along with the corresponding variance of gradient entries.}
  \label{table:ad_costs}
  \centering
  \begin{tabular}{llll}
    \toprule
    Method     & FORWARD & BACKWARD & VARIANCE \\
    \midrule
    Exact (na\"{i}ve) & $D$ $\texttt{vjp}$s  & $D$ $\texttt{vjp}$s & $0$ \\
    Exact & $d$ $\texttt{jvp}$s & $d$ $\texttt{jvp}$s & $0$  \\
    Stochastic & $K(\tau+1)$\texttt{jvp}s $+ K(\tau+1)$ \texttt{vjp}s & $K$\texttt{jvp}s $+ K$\texttt{vjp}s & $\propto 1 / K$ \\
    \bottomrule
  \end{tabular}
\end{table}

\section{Experiments}\label{sec:experiments}

We now compare our methods against the two-step baseline of \citet{brehmer2020flows}, and also study the memory vs.\ variance trade-off.
We use the real NVP \citep{dinh2016density} architecture for all flows, except we do not use batch normalization \citep{ioffe2015batch} as it causes issues with $\texttt{vjp}$ computations.
We point out that all comparisons remain fair, and we include a detailed explanation of this phenomenon in Appendix \ref{app:bn}, along with all experimental details in Appendix \ref{app:exp}.
Throughout, we use the abbreviations RNFs-ML for our maximum likelihood training method, RNFs-TS for the two-step method, and RNFs for rectangular NFs in general.
For most runs, we found it useful to anneal the likelihood term(s).
That is, at the beginning of training we optimize only the reconstruction term, and then slowly incorporate the other terms.
This likelihood annealing procedure helped avoid local optima where the manifold is not recovered (large reconstruction error) but the likelihood of projected data is high.

\subsection{Simulated Data}

\begin{figure}
    \centering
    \begin{tabular}{c c c}
    {\scriptsize RNFs-ML (exact) density} & {\scriptsize von Mises ground truth} & {\scriptsize RNFs-TS density}\\
    \includegraphics[scale=0.35]{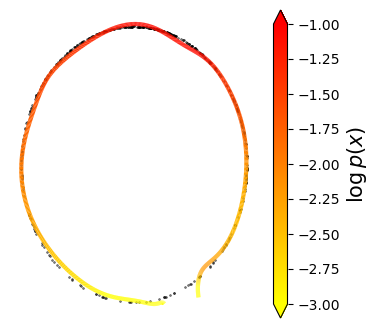} &
    \includegraphics[scale=0.35]{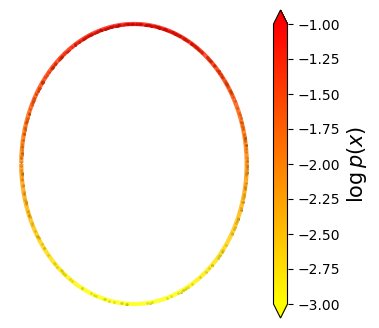} &
    \includegraphics[scale=0.35]{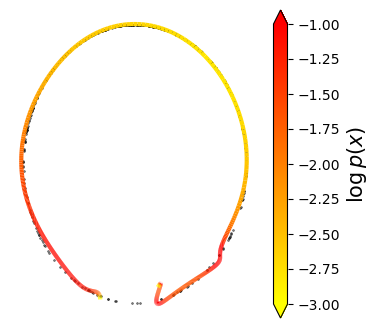}\\
    {\scriptsize RNFs-ML (exact) speed} & {\scriptsize Distribution of $f^\dagger_{\theta^*}(X)$} & {\scriptsize RNFs-TS speed}\\
     \includegraphics[trim={0 -1.5cm 0 0},clip, scale=0.35]{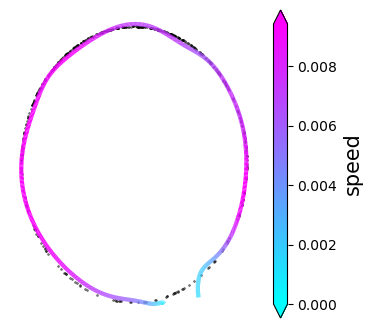} &
    \includegraphics[scale=0.35]{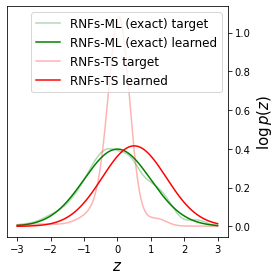} &
    \includegraphics[trim={0 -1.5cm 0 0},clip, scale=0.35]{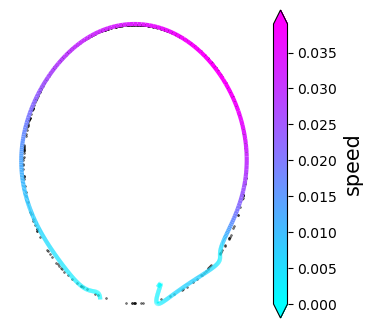}
    \end{tabular}
    \caption{Top row: RNFs-ML (exact) (left), von Mises ground truth (middle), and RNF-TS (right). Bottom row: Speed at which $f_{\theta^*}$ maps to $\mathcal{M}_{\theta^*}$ (measured as $l_2$ distance between uniformly spaced consecutive points in $\R$ mapped through $f_{\theta^*}$) for RNFs-ML (exact) (left), RNFs-TS (right), and distribution $h_\eta$ has to learn in order to recover the ground truth, fixing $\theta^*$ (middle). See text for discussion.}
    \label{fig:circle}
\end{figure}

We consider a simulated dataset where we have access to ground truth, which allows us to empirically verify the deficiencies of RNFs-TS.
We use a von Mises distribution, which is supported on the one-dimensional unit circle in $\R^2$.
\autoref{fig:circle} shows this distribution, along with its estimates from RNFs-ML (exact) and RNFs-TS.
As previously observed, RNFs-TS correctly approximate the manifold, but fail to learn the right distribution on it.
In contrast we can see that RNFs-ML, by virtue of including the Jacobian-transpose-Jacobian term in the optimization, manage to recover both the manifold and the distribution on it (\textbf{top left panel}), while also resulting in an easier-to-learn low-dimensional distribution (\textbf{bottom middle panel}) thanks to $f_{\theta^*}$ mapping to $\mathcal{M}_{\theta^*}$ at a more consistent speed (\textbf{bottom left panel}).
We do point out that, while the results presented here are representative of usual runs for both methods, we did have runs with different results which we include in Appendix \ref{app:exp} for completeness.
We finish with the observation that even though the line and the circle are not homeomorphic and thus RNFs are not perfectly able to recover the support, they manage to adequately approximate it.

\subsection{Tabular Data}

We now turn our attention to the tabular datasets used by \citet{papamakarios2017masked}, now a common benchmark for NFs as well.
As previously mentioned, one should be careful when comparing models with different supports, as we cannot rely on test likelihoods as a metric.
We take inspiration from the FID score \citep{heusel2017gans}, which is commonly used to evaluate quality of generated images when likelihoods are not available.
The FID score compares the first and second moments of a well-chosen statistic -- taken in practice to be the values of the last hidden layer of a pre-trained inception network \citep{szegedy2015going} -- from the model and data distributions using the squared Wasserstein-2 metric (between Gaussians).
Here, we take the statistic to be the data itself instead of the final hidden units of a pre-trained classifier: in other words, our metric compares the mean and covariance of generated data against those of observed data with the same squared Wasserstein-2 metric.
We include the mathematical formulas for computing both FID and our modified version for tabular data in Appendix \ref{app:fid}.
We use early stopping with our FID-like score across all models.
Our results are summarized in Table \ref{table:tabular}, where we can see that RNFs-ML consistently do a better job at recovering the underlying distribution.
Once again, these results emphasize the benefits of including the Jacobian-transpose-Jacobian in the objective.
Interestingly, except for HEPMASS, the results from our stochastic version with $K=1$ are not significantly exceeded by the exact version or using a larger value of $K$, suggesting that the added variance does not result in decreased empirical performance.
We highlight that no tuning was done (except on GAS for which we changed $d$ from $4$ to $2$), RNFs-ML outperformed RNFs-TS out-of-the-box here (details are in Appendix \ref{app:exp}).
We report training times in Appendix \ref{app:exp}, and observe that RNFs-ML take a similar amount of time as RNFs-TS to train for datasets with lower values of $D$, and while we do take longer to train for the other datasets, our training times remain reasonable and we often require fewer epochs to converge.
\begin{table}
\small
  \caption{FID-like metric for tabular data (lower is better). Bolded runs are the best or overlap with it.}
  \label{table:tabular}
  \centering
  \begin{tabular}{llllll}
    \toprule
    Method     & POWER & GAS & HEPMASS & MINIBOONE \\
    \midrule
    RNFs-ML (exact) & $\mathbf{0.067 \pm 0.016}$  & $\mathbf{0.138 \pm 0.023}$   & $\mathbf{0.486 \pm 0.032}$  & $\mathbf{0.978 \pm 0.082}$ \\
    RNFs-ML ($K=1$) & $\mathbf{0.083 \pm 0.015}$  & $\mathbf{0.110 \pm 0.021}$  & $0.779 \pm 0.191$  & $\mathbf{1.001 \pm 0.051}$  \\
    RNFs-ML ($K=10$) & $\mathbf{0.113 \pm 0.037}$  &  $\mathbf{0.140 \pm 0.013}$ & $\mathbf{0.495 \pm 0.055}$  & $\mathbf{0.878 \pm 0.083}$  \\
    RNFs-TS     & $0.178 \pm 0.024$ & $0.161 \pm 0.016$  & $0.649 \pm 0.081$  & $1.085 \pm 0.062$  \\
    \bottomrule
  \end{tabular}
\end{table}

\subsection{Image Data and Out-of-Distribution Detection}

We also compare RNFs-ML to RNFs-TS for image modelling on MNIST and FMNIST.
We point out that these datasets have ambient dimension $D=784$, and being able to fit RNFs-ML is in itself noteworthy: to the best of our knowledge no previous method has scaled optimizing the Jacobian-transpose-Jacobian term to these dimensions.
We use FID scores both for comparing models and for early stopping during training.
We also used likelihood annealing, with all experimental details again given in Appendix \ref{app:exp}.
We report FID scores in Table \ref{table:images}, where we can see that we outperform RNFs-TS.
Our RNFs-ML $(K=1)$ variant also outperforms its decreased-variance counterparts.
This initially puzzling behaviour is partially explained by the fact that we used the $K=1$ variant to tune the model (being the cheapest one to train), and then used the tuned hyperparameters for a single run of the other two variants.
Nonetheless, once again these results suggest that the variance induced by our stochastic method is not empirically harmful, and that while using the exact method should be the default whenever feasible, using $K=1$ otherwise is sensible.
We also report training times where we can see the computational benefits of our stochastic method, as well as visualizations of samples, in Appendix \ref{app:exp}.

We also compare performance on the CIFAR-10 dataset \citep{krizhevsky2009learning}, for which $D=3{,}072$.
Once again, being able to fit RNFs-ML in this setting is in itself remarkable.
We do not include RNFs-ML ($K=4$) results because of limited experimentation on CIFAR-10 due to computational cost (experimental details, including hyperparameters which we tried, are given in Appendix \ref{app:exp}).
We can see that, while RNFs-TS outperformed RNFs-ML ($K=1$) -- which we hypothesize might be reversed given more tuning -- our RNFs-ML (exact) version is the best performing model, yet again highlighting the importance of including the change-of-volume term in the objective.

\begin{wrapfigure}{r}{0.3\textwidth}
\centering
\begin{tabular}{c}
    \hspace{10pt} {\scriptsize Trained on FMNIST}  \\
    \includegraphics[width=0.27\textwidth]{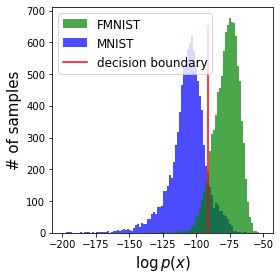}
\end{tabular}
    \caption{OoD detection with RNFs-ML (exact).}
    \label{fig:hist}
\end{wrapfigure}
We further evaluate the performance of RNFs for OoD detection. \citet{nalisnick2018deep} pointed out that square NFs trained on FMNIST assign higher likelihoods to MNIST than they do to FMNIST.
While there has been research attempting to fix this puzzling behaviour \citep{alemi2016deep, alemi2018uncertainty, choi2018waic, NEURIPS2019_1e795968}, to the best of our knowledge no method has managed to correct it using only likelihoods of trained models.
\autoref{fig:hist} shows that RNFs remedy this phenomenon, and that models trained on FMNIST assign higher test likelihoods to FMNIST than to MNIST.
This correction does not come at the cost of strange behaviour now emerging in the opposite direction (i.e.\ when training on MNIST, see Appendix \ref{app:exp} for a histogram).
\autoref{table:images} quantifies these results (arrows point from in-distribution datasets to OoD ones) with the accuracy of a decision stump using only log-likelihood, and we can see that the best-performing RNFs models essentially solve this OoD task.
While we leave a formal explanation of this result for future work, we believe this discovery highlights the importance of properly specifying models and of ensuring the use of appropriate inductive biases, in this case low intrinsic dimensionality of the observed data.
The strong performance of RNFs-TS here seems to indicate that this is a property of RNFs rather than of our ML training method specifically,
although our exact approach is still used to compute these log-likelihoods at test time.
We include additional results on OoD detection using reconstruction errors -- along with a discussion -- in Appendix \ref{app:exp}, where we found the opposite unexpected behaviour: FMNIST always has smaller reconstruction errors, regardless of which dataset was used for training.

\begin{table}
\small
  \caption{FID scores (lower is better) and decision stump OoD accuracy (higher is better).}
  \label{table:images}
  \centering
  \begin{tabular}{llllll}
    \toprule
    \multirow{2}{*}{\vspace{-4pt}Method}& \multicolumn{3}{c}{FID} & \multicolumn{2}{c}{OoD ACCURACY} \\
     \cmidrule(r){2-4} \cmidrule(r){5-6}
      & CIFAR-10 & MNIST & FMNIST & MNIST $\rightarrow$ FMNIST & FMNIST $\rightarrow$ MNIST \\
    \midrule
    RNFs-ML (exact) & $\mathbf{643.31}$ & $36.09$  & $296.01$    & $92\%$  & $91\%$  \\
    RNFs-ML ($K=1$) & $830.94$ & $\mathbf{33.98}$  & $\mathbf{288.39}$   & $97\%$  & $78\%$  \\
    RNFs-ML ($K=4$) & - & $42.90$  & $342.91$   & $77\%$  & $89\%$ \\
    RNFs-TS &   $731.46$     & $35.52$  & $318.59$  & $\mathbf{98\%}$  & $\mathbf{96\%}$   \\
    \bottomrule
  \end{tabular}
\end{table}

\section{Scope and Limitations}\label{sec:limitations}

In this paper we address the dimensionality-based misspecification of square NFs while properly using maximum likelihood as the training objective, thus providing an advancement in the training of RNFs.
Our methods however remain \emph{topologically} misspecified: even though we can better address dimensionality, we can currently only learn manifolds homeomorphic to $\R^d$.
For example, one could conceive of the MNIST manifold as consisting of $10$ connected components (one per digit), which cannot be learned by $f_\theta$.
It is nonetheless worth noting that this limitation is shared by other deep generative modelling approaches, for example GANs \citep{goodfellow2014generative} result in connected supports (since the image of a connected set under a continuous function is connected).
We observed during training in image data that the residuals of CG were not close to $\mathbf{0}$ numerically, even after $d$ steps, indicating poor conditioning and thus possible numerical non-invertibility of the matrix $J_\theta^\top J_\theta$.
We hypothesize that this phenomenon is caused by topological mismatch, which we also conjecture affects us more than the baseline as our CG-obtained (or from the exact method) gradients might point in an inaccurate direction.
We thus expect our methods in particular to benefit from improved research on making flows match the target topology, for example via continuous indexing \citep{cornish2020relaxing}.

Additionally, while we have successfully scaled likelihood-based training of RNFs far beyond current capabilities, our methods -- even the stochastic one -- remain computationally expensive for higher dimensions, and further computational gains remain an open problem.
We also attempted OoD detection on CIFAR-10 against the SVHN dataset \citep{netzer2011reading}, and found that neither RNFs-ML nor RNFs-TS has good performance, although anecdotally we may have at least improved on the situation outlined by \citet{nalisnick2018deep}.
We hypothesize these results might be either caused by topological mismatch, or corrected given more tuning.

\section{Conclusions and Broader Impact}\label{sec:conclusion}

In this paper we argue for the importance of likelihood-based training of rectangular flows, and introduce two methods allowing to do so.
We study the benefits of our methods, and empirically show that they are preferable to current alternatives.
Given the methodological nature of our contributions, we do not foresee our work having any negative ethical implications or societal consequences.

\section*{Acknowledgements}
We thank Brendan Ross, Jesse Cresswell, and Maksims Volkovs for useful comments and feedback.
We would also like to thank Rob Cornish for the excellent CIFs codebase upon which our code is built, and Emile Mathieu for plotting suggestions.
GP and JPC are supported by the Simons Foundation, McKnight Foundation, the Grossman Center, and the Gatsby Charitable Trust.

\bibliographystyle{abbrvnat}
\bibliography{non_square_flows_bib.bib}{}

\newpage
\appendix

\section{Injective Change-of-Variable Formula and Stacking Injective Flows}\label{app:change}

We first derive \eqref{eq:rectangular_density_2} from \eqref{eq:rectangular_density_proj}. By the chain rule, we have:
\begin{equation}
    \jac[g_\phi]\left(g_\phi^\dagger (x)\right) = \jac [f_\theta]\left(f_\theta^\dagger(x)\right)\jac [h_\eta]\left(g_\phi^\dagger(x)\right).
\end{equation}
The Jacobian-transpose Jacobian term in \eqref{eq:rectangular_density_proj} thus becomes:
\begin{align}\label{app_eq:decompose}
    & \left|\det \jac [g_\phi]^\top\left(g_\phi^\dagger(x)\right) \jac [g_\phi]\left(g_\phi^\dagger(x)\right)\right|^{-1/2} \\
    & = \left| \det \jac [h_\eta]^\top\left(g_\phi^\dagger(x)\right) \jac [f_\theta]^\top\left(f_\theta^\dagger(x)\right) \jac [f_\theta]\left(f_\theta^\dagger(x)\right)\jac [h_\eta]\left(g_\phi^\dagger(x)\right) \right|^{-1/2} \nonumber\\
    & = \left| \det \jac [h_\eta]^\top\left(g_\phi^\dagger(x)\right)\right|^{-1/2}\left| \det \jac [f_\theta]^\top\left(f_\theta^\dagger(x)\right) \jac [f_\theta]\left(f_\theta^\dagger(x)\right)\right|^{-1/2}\left| \det\jac [h_\eta]\left(g_\phi^\dagger(x)\right) \right|^{-1/2} \nonumber\\
    & = \left| \det \jac [h_\eta]\left(g_\phi^\dagger(x)\right)\right|^{-1}\left| \det \jac [f_\theta]^\top\left(f_\theta^\dagger(x)\right) \jac [f_\theta]\left(f_\theta^\dagger(x)\right)\right|^{-1/2} \nonumber,
\end{align}
where the second equality follows from the fact that $\jac [h_\eta]^\top(g_\phi^\dagger(x))$, $\jac [f_\theta]^\top(f_\theta^\dagger(x)) \jac [f_\theta](f_\theta^\dagger(x))$, and $\jac [h_\eta] (g_\phi^\dagger(x))$ are all square $d \times d$ matrices; and the third equality follows because determinants are invariant to transpositions. The observation that the three involved matrices are square is the reason behind why we can decompose the change-of-variable formula for $g_\phi$ as applying first the change-of-variable formula for $h_\eta$, and then applying it for $f_\theta$.

This property, unlike in the case of square flows, does not always hold. That is, the change-of-variable formula for a composition of injective transformations is not necessarily equivalent to applying the injective change-of-variable formula twice.
To see this, consider the case where $g_1 :\mathbb{R}^d\rightarrow \mathbb{R}^{d_2}$ and $g_2: \mathbb{R}^{d_2} \rightarrow \mathbb{R}^D$ are injective, where $d < d_2 < D$ and let $g = g_2 \circ g_1$. Clearly $g$ is injective by construction, and thus the determinant from its change-of-variable formula at a point $z \in \mathbb{R}^d$ is given by:
\begin{equation}\label{app_eq:correct_det}
    \det \jac [g]^\top(z) \jac [g](z) = \det \jac [g_1]^\top(z) \jac [g_2]^\top \left(g_1(z)\right) \jac [g_2]\left(g_1(z)\right) \jac [g_1](z),
\end{equation}
where now $\jac [g_1](z) \in \mathbb{R}^{d_2 \times d}$ and $\jac [g_2](g_1(z)) \in \mathbb{R}^{D \times d_2}$. Unlike the determinant from \eqref{app_eq:decompose}, this determinant cannot be easily decomposed into a product of determinants since the involved matrices are not all square. In particular, \eqref{app_eq:correct_det} need not match:
\begin{equation}
    \det \jac [g_1]^\top(z) \jac [g_1](z) \cdot \det \jac [g_2]^\top (g_1(z)) \jac [g_2] (g_1(z)),
\end{equation}
which would be the determinant terms from applying the change-of-variable formula twice. Note that this observation does not imply that a flow like $g$ could not be trained with our method, it simply implies that the $\det \jac [g]^\top (z) \jac [g](z)$ term has to be considered as a whole, and not decomposed into separate terms. It is easy to verify that in general, only an initial $d$-dimensional square flow can be separated from the overall Jacobian-transpose-Jacobian determinant.

\section{Conjugate Gradients}\label{app:cg}

We outline the CG algorithm in Algorithm \ref{app_alg:cg}, whose output we write as $\texttt{CG}(A;\epsilon)$ in the main manuscript (we omit the dependance on the tolerance $\delta$ for notational simplicity). Note that CG does not need access to $A$, just a matrix-vector product routine against $A$, $\texttt{mvp\_A}(\cdot)$. If $A$ is symmetric positive definite, then CG converges in at most $d$ steps, i.e.\ its output matches $A^{-1}\epsilon$ and the corresponding residual is $0$, and CG uses thus at most $d$ calls to $\texttt{mvp\_A}(\cdot)$. This convergence holds mathematically, but can be violated numerically if $A$ is ill-conditioned, which is why the $\tau < d$ condition is added in the while loop.

\begin{algorithm}[H]
\label{app_alg:cg}
\SetAlgoLined
\SetKwInOut{Input}{Input}
\SetKwInOut{Output}{Output}
\Input{$\texttt{mvp\_A}(\cdot)$, function for matrix-vector products against $A \in \mathbb{R}^{d\times d}$ \newline $\epsilon \in \mathbb{R}^d$ \newline $\delta \geq 0$, tolerance}
\Output{$A^{-1}\epsilon$}
$u_0 \leftarrow \mathbf{0} \in \mathbb{R}^d$ \tcp{current solution}
$r_0 \leftarrow  -\epsilon$ \tcp{current residual}
$q_0 \leftarrow r_0$\\
$\tau \leftarrow 0$\\
 \While{$||r_\tau||_2 > \delta \textbf{ and } \tau < d$}{
  $v_\tau \leftarrow \texttt{mvp\_A}(q_\tau)$\\
  $\alpha_\tau \leftarrow (r_\tau^\top r_\tau) / (q_\tau^\top v_\tau)$\\
  $u_{\tau+1} \leftarrow u_\tau + \alpha_\tau q_\tau$\\
  $r_{\tau+1} \leftarrow r_\tau - \alpha_\tau v_\tau$\\
  $\beta_\tau \leftarrow (r_{\tau+1}^\top r_{\tau+1}) / (r_\tau^\top r_\tau)$\\
  $q_{\tau+1} \leftarrow r_{\tau+1} + \beta_\tau q_\tau$\\
  $\tau \leftarrow \tau+1$\\
 }
\Return $u_\tau$
 \caption{CG}
\end{algorithm}

\section{Automatic Differentiation}\label{app:ad}

Here we summarize the relevant properties from forward- and backward-mode automatic differentiation (AD) which we use in the main manuscript. Let $f$ be the composition of smooth functions $f_1,\dots,f_L$, i.e.\ $f = f_L\circ f_{L-1} \circ \cdots \circ f_1$. For example, in our setting this function could be $f_\theta$, so that $f_1 = \texttt{pad}$, and the rest of the functions could be coupling layers from a $D$-dimensional square flow (or the functions whose compositions results in the coupling layers). By the chain rule, the Jacobian of $f$ is given by:
\begin{equation}\label{app_eq:ad}
    \jac [f](z) = \jac [f_L](\bar{f}_{L-1}(z)) \cdots \jac [f_2](\bar{f}_{1}(z)) \jac [f_1](z),
\end{equation}
where $\bar{f}_l := f_l \circ f_{l-1} \circ \cdots \circ f_1$ for $l=1,2,\dots,L-1$. Forward-mode AD computes products from right to left, and is thus efficient for computing $\texttt{jvp}$ operations. Computing $\jac [f](z) \epsilon$ is thus obtained by performing $L$ matrix-vector multiplications, one against each of the Jacobians on the right hand side of \eqref{app_eq:ad}. Backward-mode AD computes products from left to right, and would thus result in significantly more inefficient $\texttt{jvp}$ evaluations involving $L-1$ matrix-matrix products, and a single matrix-vector product. Analogously, backward-mode AD computes $\texttt{vjp}$s of the form $v^\top \jac [f](z)$ efficiently, using $L$ vector-matrix products, while forward-mode AD would require $L-1$ matrix-matrix products and a single vector-matrix product.

Typically, the cost of evaluating a matrix-vector or vector-matrix product against $\jac [f_{l+1}](\bar{f}_{l})$ (or $\jac [f_1](z)$) is the same as computing $\bar{f}_{l+1}(z)$ from $\bar{f}_l(z)$, i.e.\ the cost of evaluating $f_{l+1}$ (or the cost of evaluating $f_1$ in the case of $\jac [f_1](z)$) \citep{baydin2018automatic}. $\texttt{jvp}$ and $\texttt{vjp}$ computations thus not only have the same computational cost, but this cost is also equivalent to a forward pass, i.e.\ computing $f$.

When computing $f$, obtaining a $\texttt{jvp}$ with forward-mode AD adds the same memory cost as another computation of $f$ since intermediate results do not have to be stored. That is, in order to compute $\jac [f_l](\bar{f}_{l-1}(z)) \cdots \jac [f_1](z) \epsilon$, we only need to store $\jac [f_{l-1}](\bar{f}_{l-2}(z)) \cdots \jac [f_1](z) \epsilon$ and $\bar{f}_{l-1}(z)$ (which has to be stored anyway for computing $f$) in memory. On the other hand, computing a $\texttt{vjp}$ with backward-mode AD has a higher memory cost: One has to first compute $f$ and store all the intermediate $\bar{f}_l(z)$ (along with $z$), since computing $v^\top \jac[f_L](\bar{f}_{L-1}(z)) \cdots \jac [f_l](\bar{f}_{l-1}(z))$ from $v^\top \jac[f_L](\bar{f}_{L-1}(z)) \cdots \jac [f_{l+1}](\bar{f}_l(z))$ requires having $\bar{f}_{l-1}(z)$ in memory.

In practice, we use PyTorch's implementation of backpropagation to compute $\texttt{vjp}$s, and as mentioned in the main manuscript, we use our own implementation of forward-mode AD for $\texttt{jvp}$s.
We achieve this by having every layer and non-linearity $f_l$ in our networks not only take an input $x$, but also a vector $\epsilon$ of the same length as $x$; and not just output the usual output $f_l(x)$, but also $\jac [f_l](x) \epsilon$ (for linear layers, this is equivalent to applying the layer without the bias to $\epsilon$, and for element-wise non-linearities $\jac [f_l](x)$ is a straightforward-to-compute diagonal matrix and so $\jac [f_l](x) \epsilon$ can be obtained though element-wise products).

\section{Summary of our Proposed Methods}\label{app:summary}

We summarize our methods for computing/estimating the gradient of the log determinant arising in maximum likelihood training of rectangular flows.
Algorithm \ref{app_alg:exact} shows the exact method, where $\texttt{jvp}(f, z, \epsilon)$ denotes computing $\jac [f](z)\epsilon$ using forward-mode AD, and $\epsilon_i \in \R^d$ is the $i$-th standard basis vector, i.e.\ a one-hot vector with a $1$ on its $i$-th coordinate. Note that $\partial / \partial \theta \log \det A_\theta$ is computed using backpropagation.
The \texttt{for} loop is easily parallelized in practice.
For density evaluation, rather than returning $\partial / \partial \theta \log \det A_\theta$, the output of Algorithm \ref{app_alg:exact} becomes $\log \det A_\theta$.

Algorithm \ref{app_alg:stochastic} shows our stochastic method, where $\texttt{vjp}(f, z, v)$ denotes $v^\top \jac [f](z)$ computed through backward-mode AD.
As mentioned in the main manuscript, the $\epsilon_k$ vectors can be sampled from any zero-mean, identity-covariance distribution and not just a Gaussian.
For added clarity, we change the CG notation and use $\texttt{CG}(\texttt{mvp\_A}(\cdot), \epsilon, \delta)$ to denote the output of the conjugate gradients method.
Backpropagation is once again used to compute $\partial / \partial \theta (s_\theta / K)$, and the \texttt{for} loop is again parallelized.
Note that, unlike Algorithm \ref{app_alg:exact}, $s_\theta / K$ is not a valid log determinant estimate, and Algorithm \ref{app_alg:stochastic} should \emph{only} be used for gradient estimates during training, and not density evaluation at test time.

\begin{algorithm}[H]
\label{app_alg:exact}
\SetAlgoLined
\SetKwInOut{Input}{Input}
\SetKwInOut{Output}{Output}
\Input{$f_\theta: \R^d \rightarrow \R^D$ \newline $x \in \R^D$}
\Output{$\partial / \partial \theta \log \det J_\theta^\top J_\theta$}
$z \leftarrow f_\theta^\dagger(x)$\\
\For{$i=1,\dots,d$}{
$v_i \leftarrow \texttt{jvp}(f_\theta, z, \epsilon_i)$
}
$J_\theta \leftarrow (v_1|\dots|v_d)$\\
$A_\theta \leftarrow J_\theta^\top J_\theta$\\
\Return $\partial / \partial \theta \log \det A_\theta$
 \caption{Exact method}
\end{algorithm}

\begin{algorithm}[H]
\label{app_alg:stochastic}
\SetAlgoLined
\SetKwInOut{Input}{Input}
\SetKwInOut{Output}{Output}
\Input{$f_\theta: \R^d \rightarrow \R^D$ \newline $x \in \R^D$ \newline $K \in \mathbb{N}_+$ \newline $\delta \geq 0$, CG tolerance}
\Output{Unbiased stochastic approximation of $\partial / \partial \theta \log \det J_\theta^\top J_\theta$}
$z \leftarrow f_\theta^\dagger(x)$\\
$\texttt{mvp\_A}(\cdot) \leftarrow \texttt{vjp}(f_\theta, z, \texttt{jvp}(f_\theta, z, \cdot))^\top$\\
$s_\theta \leftarrow 0$\\
\For{$k=1,\dots,K$ }{
$\epsilon_k \sim \mathcal{N}(0, I_d)$\\
$s_\theta \leftarrow s_\theta + \texttt{stop\_gradient}(\texttt{CG}(\texttt{mvp\_A}(\cdot), \epsilon_k, \delta)^\top) \cdot \texttt{mvp\_A}(\epsilon_k)$
}
\Return $\partial / \partial \theta (s_\theta / K)$
 \caption{Stochastic method}
\end{algorithm}

\section{Batch Normalization}\label{app:bn}

We now explain the issues that arise when combining batch normalization with $\texttt{vjp}$s. These issues arise not only in our setting, but every time backward-mode AD has to be called to compute or approximate the gradient of the determinant term. We consider the case with a batch of size $2$, $x_1$ and $x_2$, as it exemplifies the issue and the notation becomes simpler. Consider applying $f_\theta$ (without batch normalization) to each element in the batch, which we denote with the batch function $F_\theta$:
\begin{equation}
    F_\theta(x_1, x_2) := \left(f_\theta(x_1), f_\theta(x_2)\right).
\end{equation}
The Jacobian of $F_\theta$ clearly has a block-diagonal structure:
\begin{equation}
    \jac [F_\theta] (x_1, x_2) = \begin{pmatrix}
\jac [f_\theta](x_1) & \mathbf{0}\\
\mathbf{0} & \jac [f_\theta](x_2)
\end{pmatrix}.
\end{equation}
This structure implies that relevant computations such as $\texttt{vjp}$s, $\texttt{jvp}$s, and determinants parallelize over the batch:
\begin{align}
    (v_1, v_2)^\top \jac [F_\theta] (x_1, x_2) & = \left(v_1^\top \jac [f_\theta](x_1), v_2^\top \jac [f_\theta](x_2)\right)\\
    \jac [F_\theta] (x_1, x_2) \begin{pmatrix}
\epsilon_1\\
\epsilon_2
\end{pmatrix} & = \begin{pmatrix}
\jac [f_\theta](x_1)\epsilon_1\\
\jac [f_\theta](x_2)\epsilon_2
\end{pmatrix}\nonumber\\
    \det \jac [F_\theta]^\top (x_1, x_2) \jac [F_\theta] (x_1, x_2) & = \det \jac [f_\theta]^\top (x_1)\jac [f_\theta] (x_1) \det \jac [f_\theta]^\top (x_2) \jac [f_\theta](x_2) \nonumber.
\end{align}
In contrast, when using batch normalization, the resulting computation $F_\theta^{BN}(x_1, x_2)$ does not have a block-diagonal Jacobian, and thus this parallelism over the batch breaks down, in other words:
\begin{align}
    (v_1, v_2)^\top \jac \left[F_\theta^{(BN)}\right] (x_1, x_2) & \neq \left(v_1^\top \jac [f_\theta](x_1), v_2^\top \jac [f_\theta](x_2)\right)\\
    \jac \left[F_\theta^{BN}\right] (x_1, x_2) \begin{pmatrix}
\epsilon_1\\
\epsilon_2
\end{pmatrix} & \neq \begin{pmatrix}
\jac [f_\theta](x_1)\epsilon_1\\
\jac [f_\theta](x_2)\epsilon_2
\end{pmatrix}\nonumber\\
    \det \jac \left[F_\theta^{BN}\right]^\top (x_1, x_2) \jac \left[F_\theta^{BN}\right] (x_1, x_2) & \neq \det \jac [f_\theta]^\top (x_1)\jac [f_\theta] (x_1) \det \jac [f_\theta]^\top (x_2) \jac [f_\theta](x_2) \nonumber,
\end{align}
where the above $\neq$ signs should be interpreted as ``not generally equal to'' rather than always not equal to, as equalities could hold coincidentally in rare cases.

In square flow implementations, AD is never used to obtain any of these quantities, and the Jacobian log determinants are explicitly computed for each element in the batch. In other words, this batch dependence is ignored in square flows, both in the log determinant computation, and when backpropagating through it. Elaborating on this point, AD is only used to backpropagate (with respect to $\theta$) over this explicit computation. If AD was used on $F_\theta^{BN}$ to construct the matrices and we then computed the corresponding log determinants, the results would not match with the explicitly computed log determinants: The latter would be equivalent to using batch normalization with a $\texttt{stop\_gradient}$ operation \textit{with respect to} $(x_1, x_2)$ \textit{but not with respect to} $\theta$, while the former would use no $\texttt{stop\_gradient}$ whatsoever. Unfortunately, this partial $\texttt{stop\_gradient}$ operation only with respect to inputs but not parameters is not available in commonly used AD libraries. While our custom implementation of $\texttt{jvp}$s can be easily ``hard-coded'' to have this behaviour, doing so for $\texttt{vjp}$s would require significant modifications to PyTorch. We note that this is \textit{not} a fundamental limitation and that these modifications could be done to obtain $\texttt{vjp}$s that behave as expected with a low-level re-implementation of batch normalization, but these fall outside of the scope of our paper. Thus, in the interest of performing computations in a manner that remains consistent with what is commonly done for square flows and that allows fair comparisons of our exact and stochastic methods, we avoid using batch normalization.

\section{FID and FID-like Scores}\label{app:fid}

For a given dataset $\{x_1,\dots, x_n\} \subset \mathbb{R}^D$ and a set of samples generated by a model $\{x_1^{(g)}, \dots, x_m^{(g)}\} \subset \mathbb{R}^D$, along with a statistic $T:\mathbb{R}^D \rightarrow \mathbb{R}^r$, the empirical means and covariances are given by:
\begin{align}
    \hat{\mu} & := \displaystyle \dfrac{1}n \sum_{i=1}^n T(x_i), \hspace{39pt}
    \hat{\Sigma} := \dfrac{1}{n-1} \displaystyle \sum_{i=1}^n \left(T(x_i) - \hat{\mu}\right)\left(T(x_i) - \hat{\mu}\right)^\top \\
    \hat{\mu}^{(g)} & := \displaystyle \dfrac{1}{m} \sum_{i=1}^m T\left(x_i^{(g)}\right), \hspace{10pt}
    \hat{\Sigma}^{(g)} := \dfrac{1}{m-1} \displaystyle \sum_{i=1}^m \left(T\left(x_i^{(g)}\right) - \hat{\mu}^{(g)}\right)\left(T\left(x_i^{(g)}\right) - \hat{\mu}^{(g)}\right)^\top.
\end{align}
The FID score takes $T$ as the last hidden layer of a pre-trained inception network, and evaluates generated sample quality by comparing generated moments against data moments. This comparison is done with the squared Wasserstein-2 distance between Gaussians with corresponding moments, which is given by:
\begin{equation}
    \left|\left|\hat{\mu} - \hat{\mu}^{(g)}\right|\right|_2^2 + \mathrm{tr}\left(\hat{\Sigma} + \hat{\Sigma}^{(g)} -2\left(\hat{\Sigma}\hat{\Sigma}^{(g)}\right)^{1/2}\right),
\end{equation}
which is $0$ if and only if the moments match. Our proposed FID-like score for tabular data is computed the exact same way, except no inception network is used. Instead, we simply take $T$ to be the identity, $T(x)=x$.

\section{Experimental Details}\label{app:exp}

First we will comment on hyperparameters/architectural choices shared across experiments.
The $D$-dimensional square flow that we use, as mentioned in the main manuscript, is a RealNVP network \citep{dinh2016density}. 
In all cases, we use the ADAM \citep{DBLP:journals/corr/KingmaB14} optimizer and train with early stopping against some validation criterion specified for each experiment separately and discussed further in each of the relevant subsections below.
We use no weight decay.
We also do not use batch normalization in any experiments for the reasons mentioned above in Appendix \ref{app:bn}.
We use a standard Gaussian on $d$ dimensions as $p_Z$ in all experiments.

\paragraph{Compute} We ran our two-dimensional experiments on a Lenovo T530 laptop with an Intel i5 processor, with negligible training time per epoch.
We ran the tabular data experiments on a variety of NVIDIA GeForce GTX GPUs on a shared cluster: we had, at varying times, access to 1080, 1080 Ti, and 2080 Ti models, but never access to more than six cards in total at once.
For the image experiments, we had access to a 32GB-configuration NVIDIA Tesla v100 GPU.
We ran each of the tabular and image experiments on a single card at a time, except for the image experiments for the RNFs-ML (exact) and ($K=10$) models which we parallelized over four cards.

\autoref{app_table:times} includes training times for all of our experiments. Since we used FID-like and FID scores for ealy stopping, we include both per-epoch and total times. Per epoch times of RNFs-ML exclude epochs where the Jacobian-transpose-Jacobian log determinant is annealed with a $0$ weight, although we include time added from this portion of training into the total time cost.
Note throughout this section we also consider one epoch of the two-step baseline procedure to be one full pass through the data training the likelihood term, and then one full pass through the data training the reconstruction term.

\begin{table}
\small
  \caption{Training times in seconds, ``$K > 1$'' means $K=10$ for tabular data and $K=4$ for images.}
  \label{app_table:times}
  \centering
  \begin{tabular}{lllllllll}
    \toprule
    \multirow{2}{*}{\vspace{-4pt}Dataset}& \multicolumn{2}{c}{RNFs-ML (exact)} & \multicolumn{2}{c}{RNFs-ML ($K=1$)} & \multicolumn{2}{c}{RNFs-ML ($K>1$)} & \multicolumn{2}{c}{RNFs-TS} \\
     \cmidrule(r){2-3} \cmidrule(r){4-5} \cmidrule(r){6-7} \cmidrule(r){8-9}
         & EPOCH & TOTAL & EPOCH & TOTAL & EPOCH & TOTAL & EPOCH & TOTAL \\
    \midrule
    POWER & $53.8$  &  $4.13\text{e}3$  & $67.4$ & $6.76\text{e}3$ & $136$  & $1.14\text{e}4$   & $45.1$  & $3.83\text{e}3$  \\
    GAS & $37.3$  & $2.51\text{e}3$   & $62.7$  & $4.51\text{e}3$ & $80.1$  & $5.24\text{e}3$    & $43.2$  & $3.49\text{e}3$ \\
    HEPMASS & $143$  & $1.01\text{e}4$   & $146$  & $8.28\text{e}3$ & $159$  & $1.20\text{e}4$    & $29.1$  & $2.42\text{e}3$\\
    MINIBOONE  & $49.3$  & $4.16\text{e}3$  & $26.3$  & $2.01\text{e}3$  & $29.8$  & $2.94\text{e}3$    & $4.61$  & $481$ \\
    MNIST  & $2.40\text{e}3$  & $2.59\text{e}5$  & $1.71\text{e}3$  & $1.57\text{e}5$  & $3.03\text{e}3$  & $3.20\text{e}5$    & $2.13e2$  & $3.90\text{e}4$ \\
    FMNIST  & $2.34\text{e}3$  & $2.59\text{e}5$  & $1.72\text{e}3$  & $1.50\text{e}5$  & $3.15\text{e}3$  & $2.10\text{e}5$    & $1.04\text{e}2$  & $1.11\text{e}4$ \\
    \bottomrule
  \end{tabular}
\end{table}

\subsection{Simulated Data}

The data for this experiment is simulated from a von Mises distribution centred at $\frac \pi 2$ projected onto a circle of radius $1$.
We randomly generate $10{,}000$ training data points and train with batch sizes of $1{,}000$.
We use $1{,}000$ points for validation, performing early stopping using the value of the full objective and halting training when we do not see any validation improvement for $50$ epochs.
We create visualizations in \autoref{fig:circle} by taking $1{,}000$ grid points equally-spaced between $-3$ and $3$ as the low-dimensional space, project these into higher dimensions by applying the flow $g_\phi$, and then assign density to these points using the injective change-of-variable formula \eqref{eq:rectangular_density}.
In this low-dimensional example, we use the full Jacobian-transpose-Jacobian which ends up just being a scalar as $d=1$.
We commence likelihood annealing (when active) on the $500$-th training epoch and end up with a full likelihood term by the $1000$-th.

For the $D$-dimensional square flow $f_\theta$, we used a $5$-layer RealNVP model, with each layer having a fully-connected coupler network of size $2 \times 10$, i.e.\ $2$ hidden layers each of size $10$, outputting the shift and (log) scale values.
The baseline additionally uses a simple shift-and-scale transformation in $d$-dimensional space as $h_\eta$; we simply use the identity map for $h_\eta$ in this simple example.

We perform slightly different parameter sweeps for the two methods based on preliminary exploration.
For the baseline two-step procedure, we perform runs over the following grid:
\begin{itemize}
    \item Learning rate: $10^{-3}$, $10^{-4}$.
    \item Regularization parameter ($\beta$): $10$, $50$, $100$, $200$, $1{,}000$, $10{,}000$ (which for this method is equivalent to having a separate learning rate for the regularization objective).
    \item Likelihood annealing: \texttt{True} or \texttt{False}.
\end{itemize}
For our method, we search over the following, although noting that our method was stable at the higher learning rate of $10^{-3}$:
\begin{itemize}
    \item Learning rate: $10^{-3}$, $10^{-4}$.
    \item Regularization parameter ($\beta$): $10$, $50$, $200$.
    \item Likelihood annealing: \texttt{True} or \texttt{False}.
\end{itemize}
Empirically we found the two-step baseline performed better with the higher regularization, which also agrees with the hyperparameter settings from their paper.
Note that we have searched over $2$ times as many runs for the baseline and still obtain better runs with our approach.

\paragraph{Divergences on RNFs-TS between our codebase and the implementation of \citet{brehmer2020flows}} Although we were able to replicate the baseline RNF-TS method, there were some different choices made in the codebase of the baseline method (available here: \url{https://github.com/johannbrehmer/manifold-flow}), which we outline below:
\begin{itemize}
    \item The baseline was trained for $120$ epochs and then selects the model with best validation score, whereas we use early stopping over an (essentially) unlimited number of epochs.
    \item The baseline weights the reconstruction term with a factor of $100$ and the likelihood term with a factor of $0.1$. This is equivalent in our codebase to setting $\beta = 1{,}000$, and lowering the learning rate by a factor of $10$.
    \item The baseline uses cosine annealing of the learning rate, which we do not use.
    \item The baseline includes a sharp Normal base distribution on the pulled-back padded coordinates. We neglected to include this as it isn't mentioned in the paper and can end up resulting in essentially a square flow construction.
    \item The baseline uses the ADAMW optimizer \citep{DBLP:conf/iclr/LoshchilovH19} to fix issues with weight decay within ADAM (which they also use). We stick with standard ADAM as we do not use weight decay.
    \item The baseline flow reparametrizes the scale $s$ of the RealNVP network as $s = \sigma(\tilde s + 2) + 10^{-3}$, where $\tilde s$ is the unconstrained scale and $\sigma$ is the sigmoid function, but this constrains the scale to be less than $1 + 10^{-3}$.
    This appears to be done for stability of the transformation (cf.\ the ResNets below). We instead use the standard parametrization of $s = \exp(\tilde s)$ as the fully-connected networks appear to be adequately stable.
    \item The Baseline uses ResNets with ReLU activation of size $2 \times 100$ as the affine coupling networks. We use MLPs with tanh activation function instead.
    \item The baseline uses a dataset which is not strictly on a manifold. The radius of a point on the circle is sampled from $\mathcal N(1, 0.01^2)$. We use a strictly one-dimensional distribution instead with a von Mises distribution on the angle as noted above.
\end{itemize}
In general, we favoured more standard and simpler choices for modelling the circle, outside of the likelihood annealing which is non-standard.

\begin{figure}[ht]
\centering

\begin{subfigure}[t]{0.32\linewidth}
\centering
\includegraphics[scale=0.45]{figs/all_2d_densities/Oct06_22-18-34_density}
\caption{LW $=$ F, $\beta = 50, \eta = 10^{-3}$}
\end{subfigure}%
\hfill
\begin{subfigure}[t]{0.32\linewidth}
\centering
\includegraphics[scale=0.45]{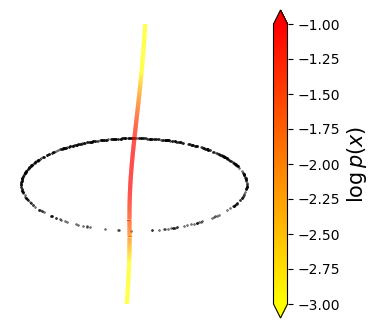}
\caption{LW $=$ F, $\beta = 200, \eta = 10^{-4}$}
\label{fig:perp-manifold}
\end{subfigure}%
\hfill
\begin{subfigure}[t]{0.32\linewidth}
\centering
\includegraphics[scale=0.45]{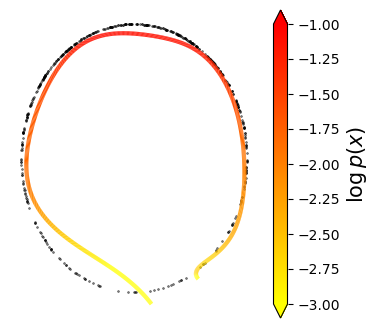}
\caption{LW $=$ F, $\beta = 10, \eta = 10^{-4}$}
\end{subfigure}%
\hfill
\begin{subfigure}[t]{0.32\linewidth}
\centering
\includegraphics[scale=0.45]{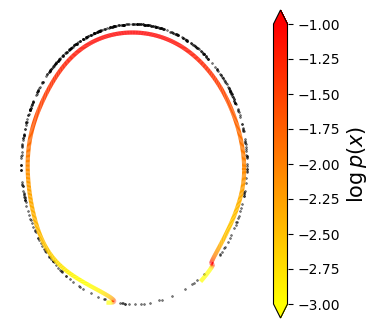}
\caption{LW $=$ T, $\beta = 10, \eta = 10^{-3}$}
\end{subfigure}%
\hfill
\begin{subfigure}[t]{0.32\linewidth}
\centering
\includegraphics[scale=0.45]{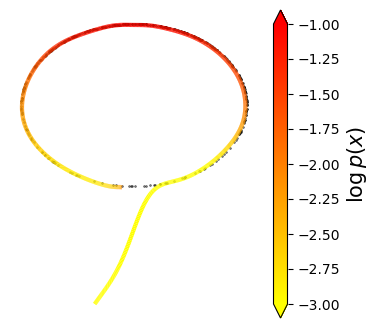}
\caption{LW $=$ F, $\beta = 50, \eta = 10^{-4}$}
\end{subfigure}%
\hfill
\begin{subfigure}[t]{0.32\linewidth}
\centering
\includegraphics[scale=0.45]{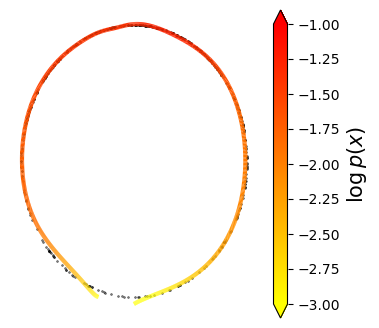}
\caption{LW $=$ F, $\beta = 200, \eta = 10^{-3}$}
\end{subfigure}%
\hfill
\begin{subfigure}[t]{0.32\linewidth}
\centering
\includegraphics[scale=0.45]{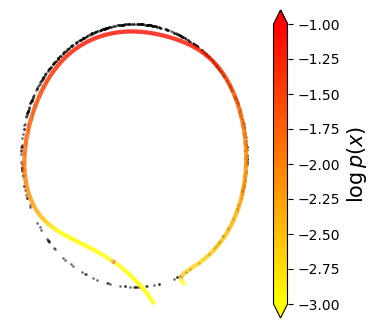}
\caption{LW $=$ T, $\beta = 10, \eta = 10^{-4}$}
\end{subfigure}%
\hfill
\begin{subfigure}[t]{0.32\linewidth}
\centering
\includegraphics[scale=0.45]{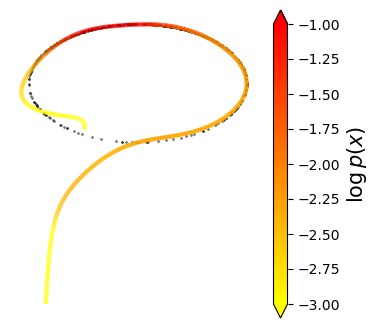}
\caption{LW $=$ T, $\beta = 200, \eta = 10^{-4}$}
\end{subfigure}%
\hfill
\begin{subfigure}[t]{0.32\linewidth}
\centering
\includegraphics[scale=0.45]{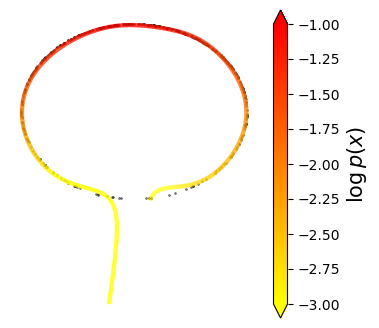}
\caption{LW $=$ T, $\beta = 50, \eta = 10^{-4}$}
\end{subfigure}%
\hfill
\begin{subfigure}[t]{0.32\linewidth}
\centering
\includegraphics[scale=0.45]{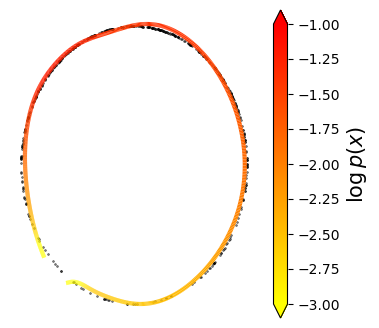}
\caption{LW $=$ T, $\beta = 50, \eta = 10^{-3}$}
\end{subfigure}%
\hfill
\begin{subfigure}[t]{0.32\linewidth}
\centering
\includegraphics[scale=0.45]{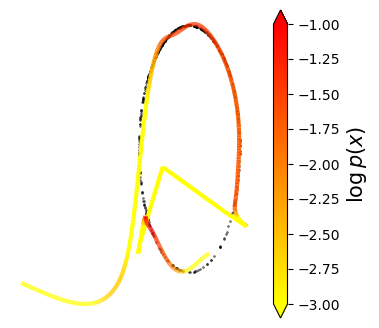}
\caption{LW $=$ T, $\beta = 200, \eta = 10^{-3}$}
\end{subfigure}%
\hfill
\begin{subfigure}[t]{0.32\linewidth}
\centering
\includegraphics[scale=0.45]{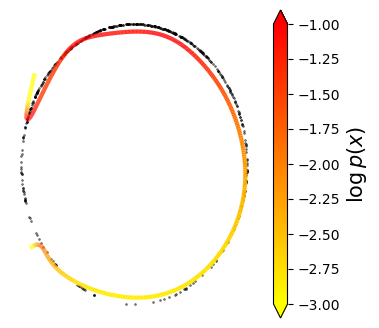}
\caption{LW $=$ F, $\beta = 10, \eta = 10^{-3}$}
\end{subfigure}%
\hfill

\caption{Runs of RNFs-ML (exact), swept over the hyperparameter combinations $\{\texttt{Likelihood Warmup} \in \{\texttt{True}, \texttt{False}\}\}\times \{\beta \in \{10, 50, 200\}\} \times \{\eta \in \{10^{-3}, 10^{-4}\}\} $}
\label{appfig:circle1}
\end{figure}

\begin{figure}[ht]
\centering

\begin{subfigure}[t]{0.32\linewidth}
\centering
\includegraphics[scale=0.45]{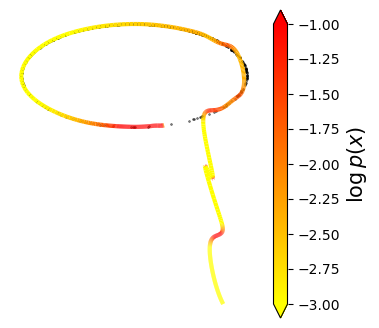}
\caption{LW $=$ T, $\beta = 1000, \eta = 10^{-3}$}
\end{subfigure}%
\hfill
\begin{subfigure}[t]{0.32\linewidth}
\centering
\includegraphics[scale=0.45]{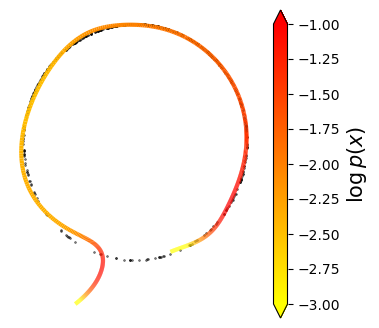}
\caption{LW $=$ T, $\beta = 100, \eta = 10^{-4}$}
\end{subfigure}%
\hfill
\begin{subfigure}[t]{0.32\linewidth}
\centering
\includegraphics[scale=0.45]{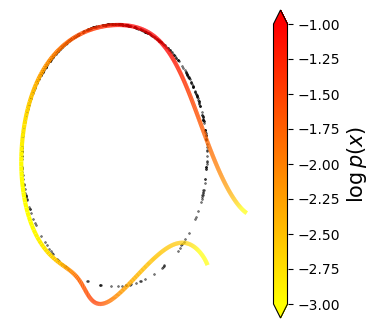}
\caption{LW $=$ F, $\beta = 10000, \eta = 10^{-3}$}
\end{subfigure}%
\hfill
\begin{subfigure}[t]{0.32\linewidth}
\centering
\includegraphics[scale=0.45]{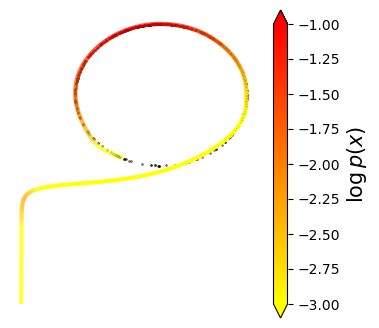}
\caption{LW $=$ T, $\beta = 10000, \eta = 10^{-3}$}
\end{subfigure}%
\hfill
\begin{subfigure}[t]{0.32\linewidth}
\centering
\includegraphics[scale=0.45]{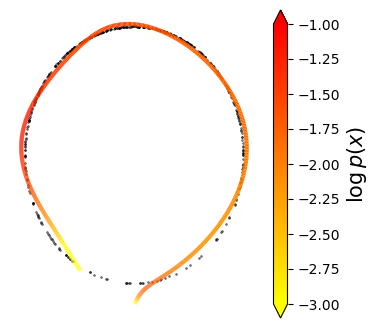}
\caption{LW $=$ T, $\beta = 100, \eta = 10^{-3}$}
\end{subfigure}%
\hfill
\begin{subfigure}[t]{0.32\linewidth}
\centering
\includegraphics[scale=0.45]{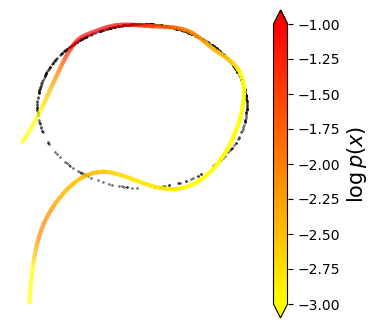}
\caption{LW $=$ F, $\beta = 100, \eta = 10^{-4}$}
\end{subfigure}%
\hfill
\begin{subfigure}[t]{0.32\linewidth}
\centering
\includegraphics[scale=0.45]{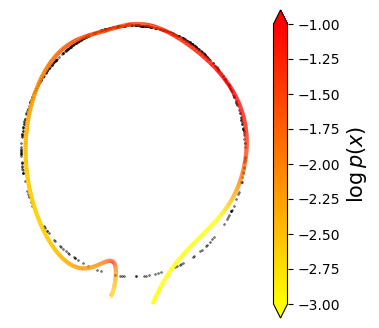}
\caption{LW $=$ F, $\beta = 100, \eta = 10^{-3}$}
\end{subfigure}%
\hfill
\begin{subfigure}[t]{0.32\linewidth}
\centering
\includegraphics[scale=0.45]{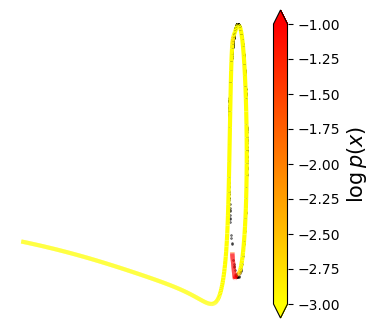}
\caption{LW $=$ F, $\beta = 1000, \eta = 10^{-3}$}
\end{subfigure}%
\hfill
\begin{subfigure}[t]{0.32\linewidth}
\centering
\includegraphics[scale=0.45]{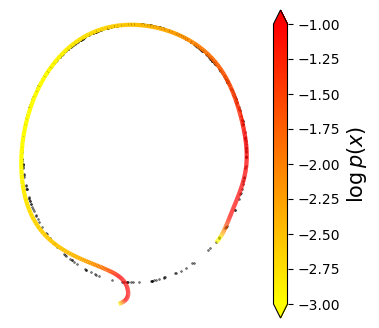}
\caption{LW $=$ T, $\beta = 10000, \eta = 10^{-4}$}
\end{subfigure}%
\hfill
\begin{subfigure}[t]{0.32\linewidth}
\centering
\includegraphics[scale=0.45]{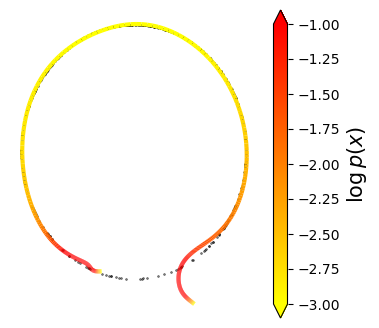}
\caption{LW $=$ F, $\beta = 1000, \eta = 10^{-4}$}
\end{subfigure}%
\hfill
\begin{subfigure}[t]{0.32\linewidth}
\centering
\includegraphics[scale=0.45]{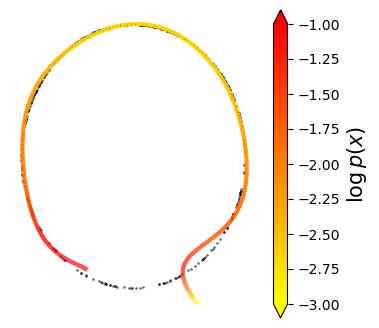}
\caption{LW $=$ T, $\beta = 1000, \eta = 10^{-4}$}
\end{subfigure}%
\hfill
\begin{subfigure}[t]{0.32\linewidth}
\centering
\includegraphics[scale=0.45]{figs/all_2d_densities/Oct06_22-34-11_density}
\caption{LW $=$ F, $\beta = 10000, \eta = 10^{-4}$}
\end{subfigure}%
\hfill

\caption{Runs of RNFs-TS, swept over the hyperparameter combinations $\{\texttt{Likelihood Warmup} \in \{\texttt{True}, \texttt{False}\}\}\times \{\beta \in \{100, 1000, 10000\}\} \times \{\eta \in \{10^{-3}, 10^{-4}\}\} $}
\label{appfig:circle2}
\end{figure}

\begin{figure}[ht]
\centering

\begin{subfigure}[t]{0.32\linewidth}
\centering
\includegraphics[scale=0.45]{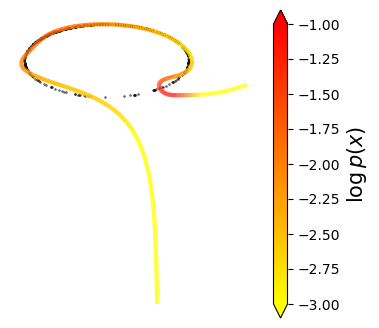}
\caption{LW $=$ F, $\beta = 200, \eta = 10^{-4}$}
\end{subfigure}%
\hfill
\begin{subfigure}[t]{0.32\linewidth}
\centering
\includegraphics[scale=0.45]{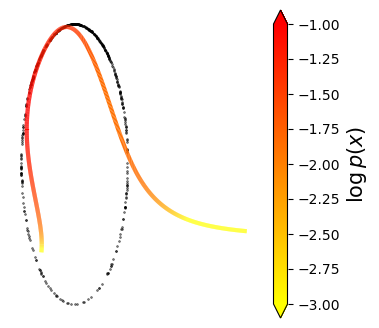}
\caption{LW $=$ F, $\beta = 10, \eta = 10^{-4}$}
\end{subfigure}%
\hfill
\begin{subfigure}[t]{0.32\linewidth}
\centering
\includegraphics[scale=0.45]{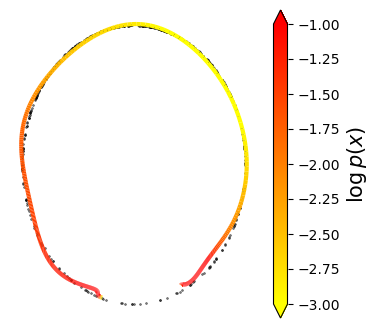}
\caption{LW $=$ T, $\beta = 200, \eta = 10^{-4}$}
\end{subfigure}%
\hfill
\begin{subfigure}[t]{0.32\linewidth}
\centering
\includegraphics[scale=0.45]{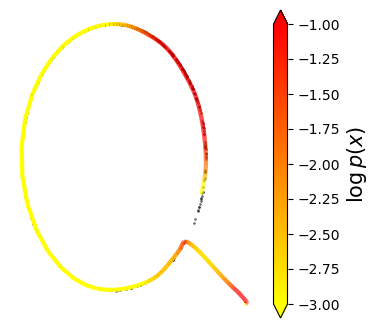}
\caption{LW $=$ T, $\beta = 50, \eta = 10^{-3}$}
\end{subfigure}%
\hfill
\begin{subfigure}[t]{0.32\linewidth}
\centering
\includegraphics[scale=0.45]{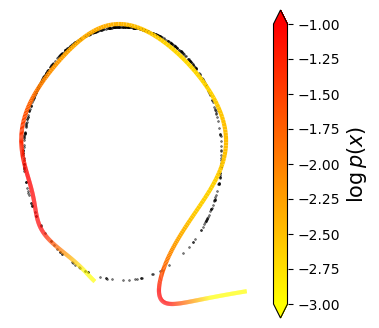}
\caption{LW $=$ F, $\beta = 50, \eta = 10^{-4}$}
\end{subfigure}%
\hfill
\begin{subfigure}[t]{0.32\linewidth}
\centering
\includegraphics[scale=0.45]{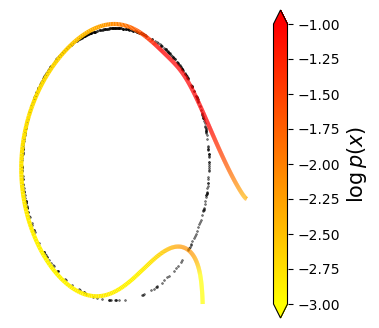}
\caption{LW $=$ F, $\beta = 10, \eta = 10^{-3}$}
\end{subfigure}%
\hfill
\begin{subfigure}[t]{0.32\linewidth}
\centering
\includegraphics[scale=0.45]{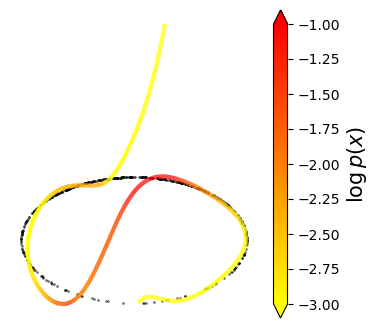}
\caption{LW $=$ F, $\beta = 200, \eta = 10^{-3}$}
\end{subfigure}%
\hfill
\begin{subfigure}[t]{0.32\linewidth}
\centering
\includegraphics[scale=0.45]{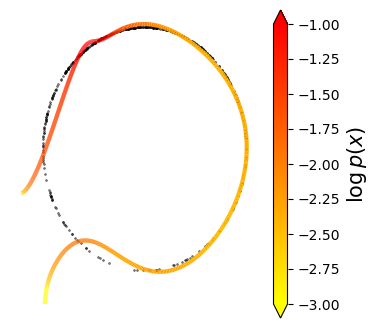}
\caption{LW $=$ T, $\beta = 50, \eta = 10^{-4}$}
\end{subfigure}%
\hfill
\begin{subfigure}[t]{0.32\linewidth}
\centering
\includegraphics[scale=0.45]{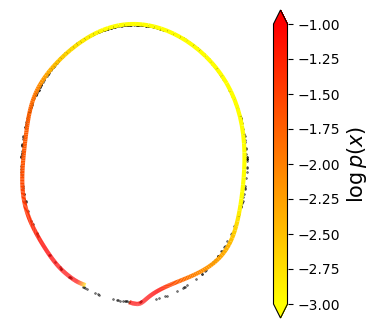}
\caption{LW $=$ T, $\beta = 10, \eta = 10^{-4}$}
\end{subfigure}%
\hfill
\begin{subfigure}[t]{0.32\linewidth}
\centering
\includegraphics[scale=0.45]{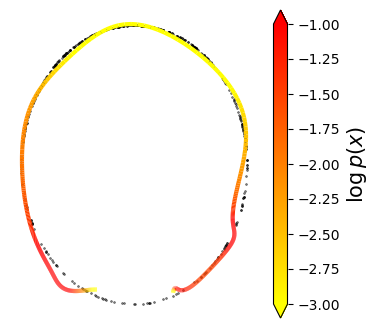}
\caption{LW $=$ F, $\beta = 50, \eta = 10^{-3}$}
\end{subfigure}%
\hfill
\begin{subfigure}[t]{0.32\linewidth}
\centering
\includegraphics[scale=0.45]{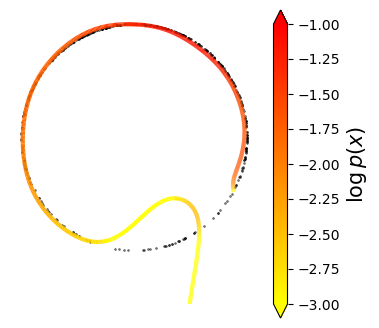}
\caption{LW $=$ T, $\beta = 200, \eta = 10^{-3}$}
\end{subfigure}%
\hfill
\begin{subfigure}[t]{0.32\linewidth}
\centering
\includegraphics[scale=0.45]{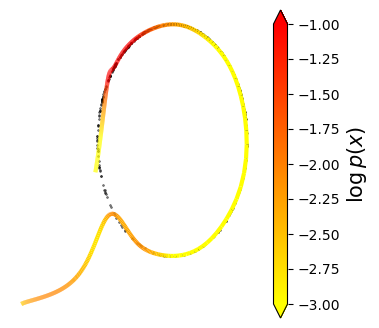}
\caption{LW $=$ T, $\beta = 10, \eta = 10^{-3}$}
\end{subfigure}%
\hfill

\caption{Runs of RNFs-TS, swept over the hyperparameter combinations $\{\texttt{Likelihood Warmup} \in \{\texttt{True}, \texttt{False}\}\}\times \{\beta \in \{10, 50, 200\}\} \times \{\eta \in \{10^{-3}, 10^{-4}\}\} $}
\label{appfig:circle3}
\end{figure}

\paragraph{Densities of \emph{all} runs} We note that, while the results reported in the main manuscript are representative of common runs, both for RNFs-ML (exact) and RNFs-TS; not every single run of RNFs-ML (exact) obtained results as good as the ones from the main manuscript. Similarly, some runs of RNFs-TS recovered better likelihoods than the one from the main manuscript. We emphasize again that the results reported on the main manuscript are the most common ones: most RNFs-ML (exact) runs correctly recovered both the manifold and the distribution on it, and most RNFs-TS runs recovered only the manifold correctly.
For completeness, we include in Figures \ref{appfig:circle1}, \ref{appfig:circle2}, and \ref{appfig:circle3} all the runs we obtained, where it becomes evident that RNFs-ML consistently outperforms RNFs-TS across runs and hyperparameter values.
However we do note in \autoref{fig:perp-manifold} that we get the perpendicular effect that \citet{brehmer2020flows} predicted might happen if optimizing the full objective, although this is far from typical of our results.

\clearpage

\subsection{Tabular Data}

For the tabular data, we use the GAS, POWER, HEPMASS, and MINIBOONE datasets, preprocessed as in \citet{papamakarios2017masked}.
We did not observe problems with overfitting in practice for any of the methods.
We use the FID-like metric with the first and second moments of the generated and observed data as described in \autoref{app:fid} for early stopping, halting training after $20$ epochs of no improvement.

We again use a RealNVP flow in $D$ dimensions but now with $10$ layers, with each layer having a fully-connected coupler network of hidden dimension $4 \times 128$.
The $d$-dimensional flow here is also a RealNVP, but just a $5$-layer network with couplers of size $2 \times 32$.

In all methods, we use a regularization parameter of $\beta = 50$.
We introduce the likelihood term with low weight after $25$ epochs, linearly increasing its contribution to the objective until it is set to its full weight after $50$ epochs.
We select $d$ as $\lfloor \frac D 2 \rfloor,$ except for ML methods on $D=8$ GAS which use $d=2$ (noted below).
We use a learning rate of $10^{-4}$.
For the methods involving the Hutchinson estimator, we use a standard Gaussian as the estimating distribution.
We also experimented with a Rademacher distribution here but found the Gaussian to be superior.

Results reported on the main manuscript are the mean of $5$ runs (with different seeds) plus/minus standard error.
Occasionally, both RNFs-ML and RNFs-TS resulted in failed runs with FID-like scores at least an order of magnitude larger than other runs.
In these rare instances, we did another run and ignored the outlier.
We did this for both methods, and we do point out that RNFs-ML did not have a higher number of failed runs.

As mentioned in the main manuscript, GAS required slightly more tuning as RNFs-ML did not outperform RNFs-TS when using $d=4$.
We instead use latent dimension $d=2$, where this time RNFs-ML did outperform. Since RNFs-TS did better with $d=4$, we report those numbers in the main manuscript.
Otherwise, our methods outperformed the baseline out-of-the-box, using parameter configurations gleaned from the image and circle experiments.

We also include the batch sizes here for completeness, which were set to be reasonably large for the purposes of speeding up the runs:
\begin{itemize}
    \item POWER - $5{,}000$
    \item GAS - $2{,}500$
    \item HEPMASS - $750$
    \item MINIBOONE - $400$
\end{itemize}

\subsection{Image Data and Out-of-Distribution Detection}

In this set of experiments, we mostly tuned the RNFs-ML methods on MNIST for $K=1$ -- applying any applicable settings to RNFs-TS on MNIST as well -- which is likely one of the main reasons that RNFs-ML perform so well for $K=1$ vs.\ the exact method or $K=4$.
The reason why we spent so much time on $K=1$ is that it was the fastest experiment to run and thus the easiest to iterate on.
Our general strategy for tuning was to stick to a base set of parameters that performed reasonably well and then try various things to improve performance.
A full grid search of all the parameters we might have wanted to try was quite prohibitive on the compute that we had available.
Some specific details on settings follow below.

For the $D$-dimensional square flow, we mainly used the $10$-layer RealNVP model which exactly mirrors the setup that \citet{dinh2016density} used on image data, except we neglect to include batch normalization (as discussed in \autoref{app:bn}) and we also tried reducing the size of the ResNet coupling networks from $8 \times 64$ to $4 \times 64$ for computational purposes.
For further computational savings, we additionally attempted to use a RealNVP with fewer layers as the $D$-dimensional square flow, but this performed extremely poorly and we did not revisit it. 
For the $d$-dimensional square component, we used another RealNVP with either $5$ or $10$ layers, and fully-connected coupler networks of size $4 \times 32$.
We also looked into modifying the flow here to be a neural spline flow \citep{durkan2019neural}, but this, like the smaller $D$-dimensional RealNVP, performed very poorly as well.
This may be because we did not constrain the norm of the gradients, although further investigation is required.
We also looked into using no $d$-dimensional flow for our methods as in the circle experiment, but this did not work well at all.

For padding, we first randomly (although this is fixed once the run begins) permute the $d$-dimensional input, pad to get to the appropriate length of vector, and then reshape to put into image dimension.
We also pad with zeros when performing the inverse of the density split operation (cf.\ the $z$ to $x$ direction of \citet[Figure~4(b)]{dinh2016density}), so that the input is actually padded twice at various steps of the flow.

When we used likelihood annealing, we did the same thing as for the tabular data: optimize only the reconstruction term for $25$ epochs, then slowly and linearly introduce the likelihood term up until it has a weight of $1$ in the objective function after epoch $50$.

We summarize our attempted parameters in \autoref{tab:params}.
For some choices of parameters, such as likelihood annealing set to \texttt{False}, $d = 15, 30$, $\beta = 10{,}000$, and CG tolerance set to $1$, we had very few runs because of computational reasons.
However, we note that the run with low CG tolerance ends up being the most successful run on MNIST. 
We have included ``SHORT NAMES'' in the table for ease of listing hyperparameter values for the runs in \autoref{table:images}, which we now provide for MNIST and FMNIST in \autoref{tab:mnist_params} and \autoref{tab:fmnist_params} respectively.
We also include batch sizes in the table.
Note that $^*$ indicates that the run was launched on $2$ GPU cards simultaneously, whereas $^{**}$ indicates that the run was launched on $4$ GPU cards.
For the CIFAR-10 parameters, we attempt several runs of the best configurations below: we sent $1$ run for RNFs-ML (exact), $2$ runs for RNFs-ML ($K=1$), and $4$ runs for RNFs-TS. 

\begin{table}[!htbp]
\centering
\caption{Parameter combinations investigated for MNIST runs. Note that the final two rows are irrelevant for RNF-ML (exact) and RNF-TS. We include "short names" for ease of listing parameters for the runs in \autoref{table:images}.}
\label{tab:params}
\begin{tabular}{l|l|ll}
\toprule
PARAMETER          & SHORT NAME     & MAIN VALUE    & ALTERNATIVES    \\
\midrule
Likelihood Annealing           & LA    & \texttt{True}          & \texttt{False}           \\
  Reconstruction parameter  &     $\beta$         & $50$          & $5, 500, 10000$ \\
Low dimension           &  $d$        & $20$          & $10,15,30$      \\
$D$-dim flow coupler    & $D$ NET    & $8 \times 64$ & $4 \times 64$   \\
$d$-dim flow layers     & $d$ LAYERS    & $5$           & $10$            \\
Hutchinson distribution & HUTCH & Gaussian      & Rademacher   \\
CG tolerance (normalized) & \texttt{tol} & $1$ & $0.001$
\end{tabular}
\end{table}

\begin{table}[!htbp]
\centering
\caption{Parameter choices for the MNIST runs reported in \autoref{table:images}.} 
\label{tab:mnist_params}
\begin{tabular}{l|llllllll}
\toprule
METHOD          & LA   & $\beta$ & $d$  & $D$ NET       & $d$ LAYERS & HUTCH    & \texttt{tol}  & BATCH  \\
\midrule
RNFs-ML (exact) & \texttt{True} & $5$     & $20$ & $8 \times 64$ & $10$       & N/A      & N/A  & $100^{**}$ \\
RNFs-ML ($K=1$) & \texttt{True} & $5$     & $20$ & $8 \times 64$ & $10$       & Gaussian & $0.001$ & $200$ \\
RNFs-ML ($K=4$) & \texttt{True} & $50$    & $20$ & $8 \times 64$ & $5$        & Gaussian & $1$   & $100^*$ \\
RNFs-TS         & \texttt{True} & $50$    & $20$ & $8 \times 64$ & $5$        & N/A      & N/A   & $200$ 
\end{tabular}
\end{table}

\begin{table}[!htbp]
\centering
\caption{Parameter choices for the FMNIST runs reported in \autoref{table:images}.} 
\label{tab:fmnist_params}
\begin{tabular}{l|llllllll}
\toprule
METHOD          & LA    & $\beta$ & $d$  & $D$ NET       & $d$ LAYERS & HUTCH      & \texttt{tol} & BATCH \\
\midrule
RNFs-ML (exact) & \texttt{True}  & $50$    & $20$ & $8 \times 64$ & $10$       & N/A        & N/A & $100^{**}$\\
RNFs-ML ($K=1$) & \texttt{True}  & $50$    & $20$ & $8 \times 64$ & $5$        & Rademacher & $1$ & $200$ \\
RNFs-ML ($K=4$) & \texttt{True}  & $50$    & $20$ & $8 \times 64$ & $10$       & Rademacher & $1$ & $200^{**}$ \\
RNFs-TS         & \texttt{False} & $5$     & $20$ & $4 \times 64$ & $10$       & N/A        & N/A  & $200$ 
\end{tabular}
\end{table}

\begin{figure}[!htbp]
    \centering
    \begin{tabular}{c c}
    \includegraphics[scale=0.55]{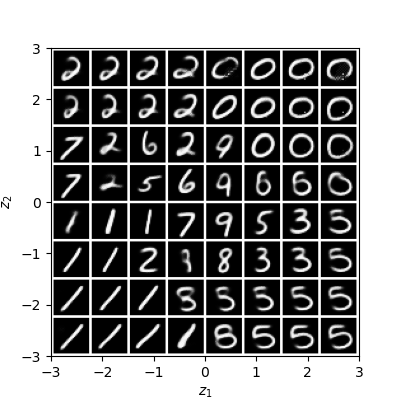} &
    \includegraphics[scale=0.55]{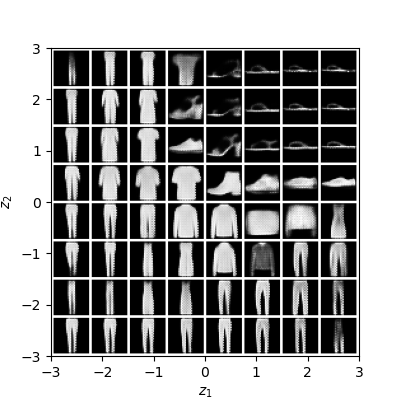}
    \end{tabular}
    \caption{MNIST (left) and FMNIST (right) samples from RNFs-ML (exact) with $d=2$. The subindices in $z_1$ and $z_2$ index coordinates, not datapoint number.}
    \label{app_fig:samples}
\end{figure}

\paragraph{Visualizations} As an attempt to visualize the learned manifold, we also trained our model with $d=2$, and show the samples obtained for different values of $z \in \R^2$ in Figure \ref{app_fig:samples}, where the spatial location of each sample is given by the Cartesian coordinates of the corresponding $z$ value.
While there are some abrupt changes, which we believe are to be expected since the true manifold likely consists of several connected components, we can see that for the most part similar-looking images have nearby latent representations.

\FloatBarrier

\begin{figure}[!htbp]
    \centering
    \begin{tabular}{c c c}
    \hspace{10pt} {\scriptsize Trained on MNIST} & \hspace{10pt} {\scriptsize Trained on FMNIST} & \hspace{10pt} {\scriptsize Trained on MNIST}\\
    \includegraphics[scale=0.35]{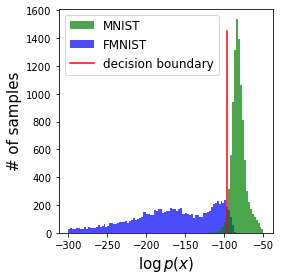} &
    \includegraphics[scale=0.35]{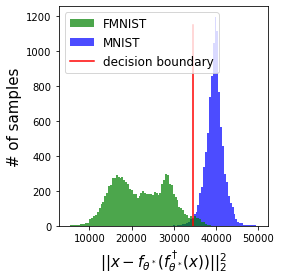} &
    \includegraphics[scale=0.35]{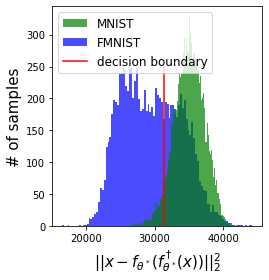}
    \end{tabular}
    \caption{OoD log-likelihood histograms trained on MNIST (left), and OoD reconstruction error histograms trained on FMNIST (middle) and MNIST (right). Log-likelihood results (left) are RNFs-ML (exact), and reconstruction results (middle and right) are RNFs-ML ($K=1$). Note that green denotes in-distribution data, and blue OoD data; and colors \emph{do not} correspond to datasets.}
    \label{app_fig:hist}
\end{figure}

\paragraph{Further Out-of-Distribution Detection Results} Figure \ref{app_fig:hist} shows RNFs-ML log-likelihoods for models trained on MNIST (\textbf{left panel}), and we can see that indeed MNIST is assigned higher likelihoods than FMNIST.
We also include OoD detection results when using reconstruction error instead of log-likelihoods, for models trained on FMNIST (\textbf{middle panel}) and MNIST (\textbf{right panel}). We observed similar results with RNFs-TS. Surprisingly, it is now the reconstruction error which exhibits puzzling behaviour: it is \emph{always} lower on FMNIST, regardless of whether the model was trained on FMNIST or MNIST.
Once again, this behaviour also happens for RNFs-TS, where the reconstruction error is optimized separately.
We thus hypothesize that this behaviour is not due to maximum likelihood training, and rather is a consequence of inductive biases of the architecture.

\end{document}